%% file: MAIN.tex
\documentclass[]{interact}
\usepackage{epstopdf}
\usepackage[caption=false]{subfig}
\usepackage[overload]{empheq}

\usepackage[numbers,sort&compress]{natbib}
\bibpunct[, ]{[}{]}{,}{n}{,}{,}
\renewcommand\bibfont{\fontsize{10}{12}\selectfont}
\makeatletter
\def\NAT@def@citea{\def\@citea{\NAT@separator}}
\makeatother

\theoremstyle{plain}

\theoremstyle{definition}

\theoremstyle{remark}

\usepackage{setspace}
\usepackage[latin1]{inputenc}
\usepackage{amsmath}
\usepackage{amsfonts}
\usepackage{graphicx}
\usepackage{subfig}
\usepackage{float}
\usepackage{siunitx}
\usepackage{listings}
\usepackage{pdfpages}
\usepackage{keyval}
\usepackage{eso-pic}
\usepackage{atbegshi}
\usepackage{pdflscape}
\usepackage{booktabs}
\usepackage[table]{xcolor}
\usepackage{multirow}
\usepackage{rotating}
\usepackage{url}
\usepackage{array}
\usepackage{fancyhdr}
\usepackage{enumitem}
\usepackage{lipsum}
\usepackage{adjustbox}
\usepackage{nicematrix}

\usepackage[np]{numprint}
\npdecimalsign{\ensuremath{.}}

\usepackage{caption}

\usepackage{algorithm,algpseudocode}
\usepackage{tikz,tikz-3dplot}
\usetikzlibrary{arrows.meta,angles,decorations.pathmorphing,patterns,patterns.meta,fadings}
\usepackage{pgfplots}
\usepackage{pgfplotstable}
\pgfplotsset{compat=1.18}
\usepgfplotslibrary{groupplots,fillbetween,statistics}
\input{Fig/TikZ_Sources/dispersion_common}

\usepackage{bm}

\usepackage{etoolbox} 

\DeclareMathOperator{\diag}{diag}

\usepackage{hyperref}
\hypersetup{%
  colorlinks=true,
  linkcolor=blue,
  citecolor=blue,
  urlcolor=blue,
  pdftitle={Parsimonious disturbance-aware minimum-time planning with parametric uncertainty},
}

\input{mymacros}

\begin{document}

\title{Parsimonious disturbance-aware minimum-time planning with parametric uncertainty}

\author{Martino Gulisano\textsuperscript{a,*}, Matteo Masoni\textsuperscript{a} and \name{Marco Gabiccini\textsuperscript{a} 
\thanks{* Contact: Martino Gulisano. Email: martino.gulisano@phd.unipi.it}
}\affil{
\textsuperscript{a} Dipartimento di Ingegneria Civile e Industriale,	Universit\`{a} di Pisa, Pisa, Italy.}
}

\maketitle

\begin{abstract}
This study presents and validates a minimum-lap-time planning (MLTP) framework for motorsport applications that embeds robustness against both state disturbances and parameter uncertainty.
The methodology builds upon a prior disturbance-aware framework that, at each track point, propagates stochastic vehicle dynamics over a short horizon and tightens tyre-friction constraints based on the worst-case scenario at horizon end. 
We extend the formulation to account for uncertainty in key vehicle parameters: moment of inertia, centre-of-mass position, and aerodynamic drag coefficient. 
To keep the extended formulation computationally tractable, a spatially selective, parsimonious activation strategy confines the robust constraints to the circuit segments where they are most critical.
We demonstrate the improved driveability of the robust references by employing a model predictive controller (MPC) as a virtual test driver.
For each reference, the same MPC drives a simulated FSAE (Formula SAE) car over 1000 runs on a representative Barcelona-Catalunya sector, with randomly realised impulsive disturbances and parameter scatter. 
We compare a nominal reference, planned without robustness, against its robust counterparts.
The latter yield consistently fewer failed runs and, at a moderate sector-time cost, show tighter dispersion of key signals (vehicle inputs, axle saturations) around the reference values, evidence of better trackability.

\end{abstract}


\input{intro}
\input{stochastic_vehicle_dynamics}
\input{robust_mltp}
\input{experiments_setup}

\input{results}
\input{conclusions}


\section*{Disclosure statement}

No potential conflict of interest was reported by the author(s).

%

\section*{Data availability statement}

The full parameter set of the vehicle model used in this paper, together
with an interactive dashboard illustrating the closed-loop Monte Carlo
results, are provided as supplementary material and openly available at \citep{Gulisano:ParsimoniousDisturbanceAwareData:2026}. The
dashboard can also be accessed directly at
\url{https://martinogulisano.github.io/data-parsimonious-disturbance-aware/}.

%
%
%
%
%
%
%

\bibliographystyle{tfnlm}
\bibliography{Disturbance-Aware-Planning}

%
%
%


%
%
%
%

\end{document}

%% file: Fig/TikZ_Sources/dispersion_common.tex
\def\datadir{post_proc/dispersion}   
\def\xcol{alpha}                     
\def\survN{1000}                     

\def\dispwidth{13cm}
\def\dispheight{4.8cm}
\def\axxmin{0.70}
\def\axxmax{0.765}   
\def\aymin{0.0}
\def\aymax{1.10}

\def\thickw{0.4pt}
\def\bandopacityinner{0.6}   
\def\bandopacityouter{0.3}   

\definecolor{nomcolor}{RGB}{0,0,0}        
\definecolor{robscolor}{RGB}{200,30,30}   
\definecolor{parscolor}{RGB}{230,160,0}   
\definecolor{parspcolor}{RGB}{0,120,200}  

\newcommand{\limitline}{%
  \addplot[gray, densely dotted, line width=0.8pt, forget plot,
           domain=\axxmin:\axxmax, samples=2] {1};%
}

\def\impulsealpha{0.725}
\newcommand{\impulseline}{%
  \draw[black, line width=0.5pt] (axis cs:\impulsealpha,\aymin) -- (axis cs:\impulsealpha,\aymax);%
}

\newcommand{\plotref}[3]{%
  \addplot[draw=none, name path=#2#1o10, forget plot]
    table[col sep=comma, x=\xcol, y=p10]{\datadir/#1/#2.csv};
  \addplot[draw=none, name path=#2#1o90, forget plot]
    table[col sep=comma, x=\xcol, y=p90]{\datadir/#1/#2.csv};
  \addplot[#3, fill opacity=\bandopacityouter, forget plot]
    fill between[of=#2#1o10 and #2#1o90];
  \addplot[draw=none, name path=#2#1i25, forget plot]
    table[col sep=comma, x=\xcol, y=p25]{\datadir/#1/#2.csv};
  \addplot[draw=none, name path=#2#1i75, forget plot]
    table[col sep=comma, x=\xcol, y=p75]{\datadir/#1/#2.csv};
  \addplot[#3, fill opacity=\bandopacityinner, forget plot]
    fill between[of=#2#1i25 and #2#1i75];
  \addplot[#3, line width=\thickw, forget plot]
    table[col sep=comma, x=\xcol, y=p50]{\datadir/#1/#2.csv};
}

\newcommand{\bandlegend}[4]{%
  \fill[white, fill opacity=0.85] (axis cs:0.7430,0.640) rectangle (axis cs:0.7635,1.06);
  \draw[gray!40, line width=0.2pt]  (axis cs:0.7430,0.640) rectangle (axis cs:0.7635,1.06);
  \fill[#1, fill opacity=\bandopacityouter] (axis cs:0.7450,0.660) rectangle (axis cs:0.7475,0.980);
  \fill[#1, fill opacity=\bandopacityinner] (axis cs:0.7450,0.746) rectangle (axis cs:0.7475,0.894);
  \draw[#1, line width=\thickw] (axis cs:0.7450,0.820) -- (axis cs:0.7475,0.820);
  \node[anchor=south, font=\tiny, inner sep=1pt] at (axis cs:0.74625,0.985) {#2};
  \fill[#3, fill opacity=\bandopacityouter] (axis cs:0.7515,0.660) rectangle (axis cs:0.7540,0.980);
  \fill[#3, fill opacity=\bandopacityinner] (axis cs:0.7515,0.746) rectangle (axis cs:0.7540,0.894);
  \draw[#3, line width=\thickw] (axis cs:0.7515,0.820) -- (axis cs:0.7540,0.820);
  \node[anchor=south, font=\tiny, inner sep=1pt] at (axis cs:0.75275,0.985) {#4};
  \draw[->, black, line width=0.25pt] (axis cs:0.7558,0.950) -- (axis cs:0.7540,0.930);
  \node[anchor=west, font=\tiny, inner sep=1pt] at (axis cs:0.7558,0.950) {10\textsuperscript{th}--90\textsuperscript{th}};
  \draw[->, black, line width=0.25pt] (axis cs:0.7558,0.820) -- (axis cs:0.7540,0.820);
  \node[anchor=west, font=\tiny, inner sep=1pt] at (axis cs:0.7558,0.820) {median};
  \draw[->, black, line width=0.25pt] (axis cs:0.7558,0.690) -- (axis cs:0.7540,0.780);
  \node[anchor=west, font=\tiny, inner sep=1pt] at (axis cs:0.7558,0.690) {25\textsuperscript{th}--75\textsuperscript{th}};
}

%% file: mymacros.tex
\newcommand{\Si}{\Sigma}

\newcommand{\Om}{\Omega}

\newcommand{\al}{\alpha}
\newcommand{\be}{\beta}

\newcommand{\ga}{\gamma}
\newcommand{\de}{\delta}

\newcommand{\sig}{\sigma}

\newcommand{\bsig}{\boldsymbol{\sig}}  
\newcommand{\bSi}{\boldsymbol{\Si}}
\newcommand{\bOm}{\boldsymbol{\Om}}  
\newcommand{\bxi}{\boldsymbol{\xi}}  
\newcommand{\bmu}{\boldsymbol{\mu}}
\newcommand{\bPhi}{\boldsymbol{\Phi}}

\newcommand{\bPsi}{\boldsymbol{\Psi}}
\newcommand{\bPsimu}{\bPsi^{\mu}}%
\newcommand{\dbmu}{\dot{\bmu}}

\newcommand{\bp}{\boldsymbol{p}}

\newcommand{\bu}{\boldsymbol{u}}

\newcommand{\bw}{\boldsymbol{w}}
\newcommand{\bx}{\boldsymbol{x}}     

\newcommand{\bz}{\boldsymbol{z}}

\newcommand{\bA}{\boldsymbol{A}}     

\newcommand{\bI}{\boldsymbol{I}}

\newcommand{\bP}{\boldsymbol{P}}
\newcommand{\bQ}{\boldsymbol{Q}}

\newcommand{\bzero}{\boldsymbol{0}}

\newcommand{\dbP}{\dot{\bP}}

\newcommand{\barh}{{\bar{h}}}

\newcommand{\tq}{{\tilde{q}}}

\newcommand{\bbR}{\mathbb{R}}    

\newcommand{\calN}{\mathcal{N}}

\newcommand{\calI}{\mathcal{I}}

\DeclareMathOperator{\Prob}{Pr}

\newcommand{\dd}{\text{{\sl d}}}        

\newcommand{\na}{\nabla}

\def\pd{\partial}

\newcommand{\p}{^{\!\cdot\!}}  

\newcommand{\st}{\bx}
\newcommand{\pa}{\bp}
\newcommand{\aug}{\boldsymbol{\eta}}
\newcommand{\dst}{\dot{\st}}
\newcommand{\dpa}{\dot{\pa}}
\newcommand{\daug}{\dot{\aug}}

\newcommand{\STM}{\bPhi}
\newcommand{\dSTM}{\dot{\STM}}
\newcommand{\bPsiSTM}{\bPsi^{\text{$\Phi$}}}%

\newcommand{\calC}{\mathcal{C}}                  
\newcommand{\calCsat}{\calC_{\mathrm{sat}}}      
\newcommand{\calCthr}{\calC_{\mathrm{thr}}}      
\newcommand{\calCpar}{\calC_{\mathrm{par}}}      

%% file: intro.tex
\section{Introduction} \label{sec:intro}

Minimum-lap-time planning (MLTP) is a standard tool in motorsport engineering: by solving a constrained optimal control problem (OCP), it synthesises the time-optimal trajectory and control inputs for a given vehicle racing on a circuit.
Engineers use these references offline, to compare vehicle setups and race strategies, and online, as feedforward signals for driver-assistance and autonomous-racing systems.
In this context, \emph{optimality} typically refers to lap-time minimisation. Since MLTP drives the vehicle as fast as the dynamic model and the OCP constraints allow, the resulting trajectory necessarily skirts the tyre-friction limit with essentially no margin to spare.

A trajectory with no margin is a fragile one to follow.
Any disturbance---such as a kerb strike or vehicle parameters that deviate from their nominal values---can push the vehicle past the limit the reference assumed.
The resulting reference is then hard, or even unsafe, to reproduce, whether followed by an expert driver or a tracking controller.

We address that weakness by planning offline robust minimum-lap-time references. 
Lap time remains the objective, but we tighten tyre-friction limits probabilistically so that the robust reference carries explicit safety margins precisely where they are most beneficial along the track.
Most MLTP formulations do not pursue that goal directly: many works adopt heuristic conservative margins or different assumptions on performance envelopes rather than embedding probabilistic robustness in the optimal control problem itself. Even the frameworks that do embed it consider uncertainty in the vehicle state alone, leaving the vehicle's own parameters fixed.
We close this gap by extending disturbance-aware minimum-lap-time planning to joint state and parameter uncertainty, providing an efficient parsimonious strategy to offset the increased computational burden, and assessing the robust MLTP references through a Monte Carlo validation campaign.

\subsection{Related work}

Minimum-lap-time planning spans a broad and well-surveyed landscape~\cite{Massaro:MinimumlaptimeOptimisationSimulation:2021}.
Quasi-steady-state formulations keep the state space small and push vehicle complexity into performance-envelope constraints~\cite{Veneri:FreetrajectoryQuasisteadystateOptimalcontrol:2020, Lovato:ThreedimensionalFreetrajectoryQuasisteadystate:2022, Biniewicz:QuasisteadystateMinimumLap:2024}, whereas high-fidelity multibody models trade computation for accuracy on three-dimensional circuits~\cite{Domenighini:MinimumLapTimePlanningMultibody:2023, Bartali:ReducedorderLieGroupbased:2023}.
Orthogonal efforts speed up these deterministic problems through better OCP formulation~\cite{DalBianco:ComparisonDirectIndirect:2019, Bertolazzi:DirectIndirectApproach:2025} or, as the program grows, through parallelization strategies~\cite{Bartali:SchwarzDecompositionParallel:2024, Bartali:ConsensusbasedAlternatingDirection:2024}.
In contrast to these offline-planning methods, a second branch of research adopts online controllers to recover much of the gap to an offline-optimal reference on a comparable high-fidelity model~\cite{Piccinini:HowOptimalMinimumtime:2024}.
Yet across all these studies the MLTP stays nominal.

A smaller group of methods instead accounts for disturbances within planning, trading some lap time for a reference easier to follow under perturbation.
Among online-planning studies, \cite{Timings:RobustLaptimeSimulation:2014a} plans a nominal lap-time trajectory and then adds a feedback controller that rejects disturbances inferred from a road-roughness model, reducing the steering effort needed to track the reference, thereby enhancing its \emph{driveability}.
The offline-planning framework~\cite{Gulisano:DisturbanceawareMinimumtimePlanning:2025} instead embeds robustness inside a single large-scale MLTP, modelling state uncertainty and propagating the state covariance. That framework enforces a probabilistic back-off on the OCP constraints, tightening the stay-on-track and friction-limit constraints.
Those frameworks, however, assume nominal values for the vehicle's parameters.

Outside motorsport, trajectory optimisation under parametric uncertainty is a far more developed topic, e.g., in robotics and control applications.
The dominant approach minimises the trajectory's sensitivity to the uncertain parameters as a term in the cost~\cite{Ansari:MinimumSensitivityControl:2016}, an idea refined from open-loop to closed-loop sensitivity~\cite{Giordano:TrajectoryGenerationMinimum:2018} and later extended to the control inputs~\cite{Brault:RobustTrajectoryPlanning:2021}.
Treating robustness as a cost penalty suits objectives that tolerate leaving the nominal optimum for a less sensitive one.
Minimum-time planning does not: its optimum lies on the friction limit, so penalising sensitivity only shifts the nominal point along that limit without building in any margin.
Under a real perturbation, the trajectory then meets the limit with nothing in reserve.
The penalty weight, moreover, fixes no probability of respecting the limit.
We therefore keep the same modelling stance---key vehicle parameters treated as uncertain---but move robustness from the cost into the constraints, tightening the friction limit by a chance-constrained back-off sized to a chosen probability of staying within it, and costing nothing where that limit is slack.

Extending robustness to parameters is not free.
Propagating the joint state-and-parameter covariance and tightening the OCP constraints against it enlarges the nonlinear program markedly.
Parsimonious strategies, however, may contain this cost. 
Concentrating the computational effort where the problem is locally hardest is the principle that drives adaptive refinement in numerical optimal control~\cite{Bartali:PnhAdaptiveRefinementProcedure:2023}.
Here, we activate the robust machinery only on critical and near-critical track regions, with an original method based on Lagrange multipliers and residuals of the OCP constraints.

Finally, assessing how well a robust reference survives execution inherently requires closed-loop evaluation. Validating disturbance-aware references from the same planning framework with professional drivers on a dynamic driving simulator offers realistic but sparse evidence~\cite{Masoni:DisturbanceAwareMinimumTimeTrajectory:2025}---a handful of runs per condition. 
Model predictive control (MPC) closes the loop computationally instead, tracking a planned reference in the presence of disturbances, and Monte Carlo campaigns are a standard way to probe a tracking controller's robustness under uncertainty~\cite{Vaskov:FrictionadaptiveStochasticNonlinear:2024, Meinert:RobustSpacecraftAttitude:2024, Ma:ModelandDatadrivenPredictive:2023}; in those studies, the campaign certifies the controller itself.
We take the complementary route: a nonlinear MPC acts as a virtual test driver tracking externally planned references in a Monte Carlo campaign. Survival rates and control effort then become quantitative measures of driveability, and lap-time increase quantifies the cost of robustness.

\subsection{Paper's contributions and organisation}

The paper makes three contributions.
\begin{itemize}
\item \textbf{Parametric uncertainty.} We extend the prior disturbance-aware MLTP framework~\cite{Gulisano:DisturbanceawareMinimumtimePlanning:2025} to propagate Gaussian uncertainty in the yaw moment of inertia, centre-of-mass position, and aerodynamic drag coefficient within the same probabilistic back-off as the state disturbances.
\item \textbf{Parsimonious activation.} We confine the robust machinery to critical and near-critical nodes, recovering the full-grid-enforcement trajectory at a fraction of the solve time.
\item \textbf{MPC Monte Carlo validation.} An MPC virtual driver tracks planned MLTP references over 1000 runs each, showing that the robust references satisfy the friction-limit constraints more reliably, at a moderate lap-time cost and with lower steering effort.
\end{itemize}

The remainder of the paper is organised as follows.
Section~\ref{sec:stochastic_veh_dyn} introduces the stochastic vehicle-dynamics framework: the single-track model with nonlinear tyres, the continuous- and discrete-time planning problems, and the probabilistic back-off used to tighten the friction-limit constraint.
Section~\ref{sec:robust_mltp} formulates the resulting open-loop covariance-propagation planning problem and its parsimonious variant.
Section~\ref{sec:experiments_setup} describes the experimental setup---reference planning, MPC design, and the Monte Carlo testing strategy---while Section~\ref{sec:results} presents and discusses the results.
Section~\ref{sec:conclusions} concludes and outlines directions for future work.

%% file: stochastic_vehicle_dynamics.tex
\section{Handling model uncertainty} \label{sec:stochastic_veh_dyn}

\subsection{Stochastic vehicle model} \label{subsec:vehicle_model}
\begin{figure}
	\centering
	\scalebox{0.85}{\input{Fig/TikZ_Sources/tikz_vehicle_topview.tex}}
	\caption{Single Track model.}
	\label{fig:single_track}
\end{figure}
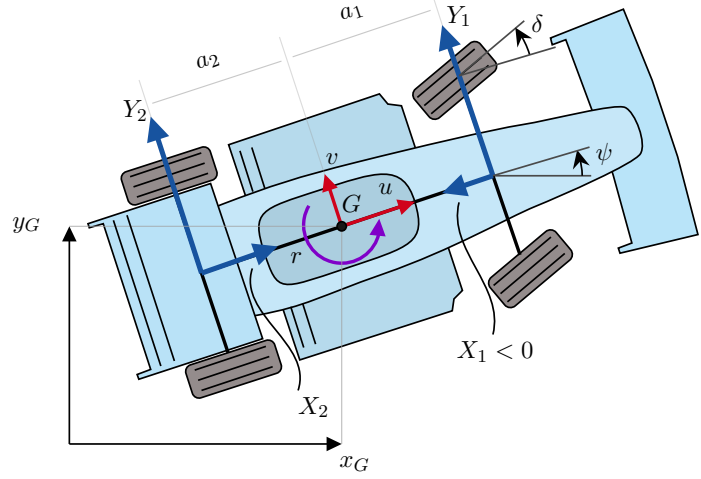

We adopt a Single Track model~\cite{Guiggiani:ScienceVehicleDynamics:2023} with nonlinear axle characteristics, illustrated in Figure~\ref{fig:single_track}.
The state vector $\st$, with $n_x=6$ components, comprises the longitudinal and lateral velocities $u$ and $v$ of the centre of mass (CoM) with respect to its body-fixed frame, the yaw rate $r$, the planar position $(x_G, y_G)$, and the heading $\psi$ (Figure~\ref{fig:single_track}).
Four uncertain parameters---the yaw moment of inertia $J_z$, the CoM height $h$, the weight balance $w_b = \frac{a_2}{a_1+a_2}$ (with $a_1$, $a_2$ the distances from the CoM to the front and rear axles), and the longitudinal drag coefficient $C_x$---form the parameter vector $\pa$, with $n_p=4$. Table~\ref{tab:vehicle_params} reports the nominal values of the main vehicle parameters; hereafter, \emph{parameters} refers exclusively to these four uncertain quantities (shaded in Table~\ref{tab:vehicle_params}), for which the listed values are the distribution means. The full parameter set is available as supplementary material~\citep{Gulisano:ParsimoniousDisturbanceAwareData:2026}.
\begin{table}
	\centering
	\setlength{\tabcolsep}{8pt}
	\caption{Nominal vehicle parameters; shaded rows denote the four uncertain quantities, with the listed values as their distribution means.}
	\label{tab:vehicle_params}
	\begin{tabular}{lrl}
		\toprule
		Symbol & Value & Description \\
		\midrule
		$m$                    & $300$\,kg            & total mass \\
		\rowcolor{gray!15}
		$J_z$                  & $157$\,kg\,m$^2$     & yaw moment of inertia \\
		$l$                    & $1.53$\,m            & wheelbase $a_1{+}a_2$ \\
		\rowcolor{gray!15}
		$w_b$                  & $0.42$               & weight balance $a_2/(a_1{+}a_2)$ \\
		\rowcolor{gray!15}
		$h$                    & $0.30$\,m            & CoM height \\
		\rowcolor{gray!15}
		$C_x$                  & $0.84$               & aerodynamic drag coefficient \\
		$C_{z1}$, $C_{z2}$     & $0.536$,\ $0.804$    & front/rear downforce coefficients \\
		$X_1/X_2$              & $0.6/0.4$            & brake balance \\
		\bottomrule
	\end{tabular}
\end{table}
Tracking the joint state-parameter distribution requires a single covariance matrix; we therefore concatenate $\st$ and $\pa$ into the augmented state $\aug = [\st;\pa]$, with $n_\eta = n_x + n_p$ components.
The control inputs are the total longitudinal force $X = X_1+X_2$ and the front steering angle $\de$, collected in the input vector $\bu$, with $n_u=2$.

The rear-wheel-drive configuration of the modelled FSAE vehicle implies that a positive (driving) longitudinal force $X$ acts only on the rear axle, while a negative (braking) force is distributed between the front and rear axles, according to a fixed brake balance~$\frac{X_1}{X_2}$.
Since the input and the fixed brake balance assign the longitudinal forces $X_1$ and $X_2$, the computation of vertical load transfers is explicit and allows direct evaluation of the lateral forces $Y_1$ and $Y_2$ using Pacejka's Magic Formula. 

The \emph{perturbed vehicle dynamics} thus takes the explicit form
\begin{equation}\label{eq:dynamics}
	\daug(t) = f(\aug(t),\bu(t))+\bw(t),
\end{equation}
where $\bw(t)$ is additive Gaussian white noise with zero mean and known covariance $\bQ(t)$, i.e., $\bw(t) \sim \calN(\bzero, \bQ(t))$.
Assuming an initially Gaussian distribution on $\aug$ and linearising the covariance dynamics to first order, the distribution remains Gaussian throughout, with mean $\bmu(t)$ and covariance $\bP(t)$, i.e., $\aug(t) \sim \calN(\bmu(t), \bP(t))$.

Since the vehicle parameters are uncertain but constant, Equation~\eqref{eq:dynamics} can be expanded into
\begin{align} \label{eq:expanded_dyn}
	\begin{bmatrix}
		\dst(t) \\[0.25em]
		\dpa(t)
	\end{bmatrix}
	=
	\begin{bmatrix}
		f_x(\st(t),\pa(t),\bu(t)) \\[0.25em]
		\mathbf{0}
	\end{bmatrix}
	+
	\begin{bmatrix}
		\bw_x(t) \\[0.25em]
		\mathbf{0}
	\end{bmatrix}.
\end{align}

Since the distribution of $\aug(t)$ remains Gaussian, the mean $\bmu(t)$ and the covariance $\bP(t)$ fully characterise its time evolution.

The mean $\bmu(t)$ follows the deterministic dynamics given by
\begin{equation}\label{eq:mean_dyn}
\dbmu(t) = f(\bmu(t), \bu(t)).
\end{equation}
The covariance $\bP(t)$ evolves according to the \emph{Lyapunov matrix differential equation}~\cite{Gajic:LyapunovMatrixEquation:1995}
\begin{align}\label{eq:dP}
\dbP(t) = \bA(t) \bP(t) + \bP(t) \bA^T(t) + \bQ(t),\quad \bP(t_0) = \bP_0 = \bP_0^T.
\end{align}
In Equation~\eqref{eq:dP}, $\bP_0$ is the \emph{initial} augmented state covariance and $\bA(t)$ is the usual shorthand notation for the Jacobian along the mean trajectory $\bmu(t)$, that is $\bA(t)=\frac{\pd f(\bmu, \bu)}{\pd \bmu}$.

To gain insight into~\eqref{eq:dP}, we show how a diagonal $\bP_0$---with state and parameter uncertainties assumed independent---propagates, by partitioning the matrices conformally with $\aug = [\st;\pa]$ and shading the non-zero blocks:
\begin{align}\label{eq:dP_color1}
	\dbP(0)
	=
	\underbrace{%
	\begin{bNiceMatrix}[margin,
		columns-width = 0.4em,
		cell-space-top-limit = 0.4em,
		cell-space-bottom-limit = 0.4em]
		\Block[fill=gray!25, tikz={pattern={Lines[angle=45,distance=6pt,line width=0.3pt]}, pattern color=black}]{2-3}{}  &  &  \\
									&  &                             \\
		\hline
		\Block[fill=white]{1-3}{}    &  &
	\end{bNiceMatrix}
	}_{\displaystyle \bA(0)}
	\underbrace{%
	\begin{bNiceMatrix}[margin,vlines=3,
		columns-width = 0.4em,
		cell-space-top-limit = 0.4em,
		cell-space-bottom-limit = 0.4em]
		\Block[fill=red!25]{2-2}{}  &  &  \\
									&  &  \\
		\hline
									&  & \Block[fill=blue!25]{1-1}{} 
	\end{bNiceMatrix}
	}_{\displaystyle \bP_0}
	+
	\underbrace{%
	\begin{bNiceMatrix}[margin,vlines=3,
		columns-width = 0.4em,
		cell-space-top-limit = 0.4em,
		cell-space-bottom-limit = 0.4em]
		\Block[fill=red!25]{2-2}{}  &  &  \\
									&  &  \\
		\hline
									&  & \Block[fill=blue!25]{1-1}{} 
	\end{bNiceMatrix}
	}_{\displaystyle \bP_0}
	\underbrace{%
	\begin{bNiceMatrix}[margin,vlines=3,
		columns-width = 0.4em,
		cell-space-top-limit = 0.4em,
		cell-space-bottom-limit = 0.4em]
		\Block[fill=gray!25, tikz={pattern={Lines[angle=45,distance=6pt,line width=0.3pt]}, pattern color=black}]{3-2}{}  &   & \Block[fill=white]{3-1}{}  \\
								   &   &  \\
								   &   &  
	\end{bNiceMatrix}
	}_{\displaystyle \bA^T(0)}
	+
	\underbrace{%
	\begin{bNiceMatrix}[margin,vlines=3,
		columns-width = 0.4em,
		cell-space-top-limit = 0.4em,
		cell-space-bottom-limit = 0.4em]
		\Block[fill=gray!25, tikz={pattern={Lines[angle=135,distance=6pt,line width=0.3pt]}, pattern color=black}]{2-2}{}  &  & \Block[fill=white]{2-1}{} \\
									&  &                           \\
		\hline
		\Block[fill=white]{1-2}{}    &  & \Block[fill=white]{1-1}{}                      
	\end{bNiceMatrix}
	}_{\displaystyle \bQ}
\end{align}
The red and blue blocks represent the initial state and parameter uncertainty, respectively.
The resulting structure of $\dbP(0)$,
\begin{align}\label{eq:dP_color2}
	\underbrace{%
	\begin{bNiceMatrix}[margin,vlines=3,
		columns-width = 0.4em,
		cell-space-top-limit = 0.4em,
		cell-space-bottom-limit = 0.4em]
		\Block[fill=red!32, tikz={pattern={Lines[angle=45,distance=6pt,line width=0.3pt]}, pattern color=black}]{2-2}{}  &  & \Block[fill=blue!32, tikz={pattern={Lines[angle=45,distance=6pt,line width=0.3pt]}, pattern color=black}]{2-1}{} \\
									&  &  \\
		\hline
									&  &  
	\end{bNiceMatrix}
	}_{\displaystyle \bA(0) \bP_0}
	+
	\underbrace{%
	\begin{bNiceMatrix}[margin,vlines=3,
		columns-width = 0.4em,
		cell-space-top-limit = 0.4em,
		cell-space-bottom-limit = 0.4em]
		\Block[fill=red!32, tikz={pattern={Lines[angle=45,distance=6pt,line width=0.3pt]}, pattern color=black}]{2-2}{}  &  &  \\
									&  &  \\
		\hline
		\Block[fill=blue!32, tikz={pattern={Lines[angle=45,distance=6pt,line width=0.3pt]}, pattern color=black}]{1-2}{} &  &  
	\end{bNiceMatrix}
	}_{\displaystyle \bP_0 \bA^T(0)}
	+
	\underbrace{%
	\begin{bNiceMatrix}[margin,vlines=3,
		columns-width = 0.4em,
		cell-space-top-limit = 0.4em,
		cell-space-bottom-limit = 0.4em]
		\Block[fill=gray!25, tikz={pattern={Lines[angle=135,distance=6pt,line width=0.3pt]}, pattern color=black}]{2-2}{}  &  & \Block[fill=white]{2-1}{} \\
									&  &                           \\
		\hline
		\Block[fill=white]{1-2}{}    &  & \Block[fill=white]{1-1}{}                      
	\end{bNiceMatrix}
	}_{\displaystyle \bQ}
	=
	\underbrace{%
	\begin{bNiceMatrix}[margin,vlines=3,
		columns-width = 0.4em,
		cell-space-top-limit = 0.4em,
		cell-space-bottom-limit = 0.4em]
		\Block[fill=red!40, tikz={postaction={pattern={Lines[angle=45,distance=6pt,line width=0.3pt]}, pattern color=black}, postaction={pattern={Lines[angle=-45,distance=6pt,line width=0.3pt]}, pattern color=black}}]{2-2}{}  &  & \Block[fill=blue!32, tikz={pattern={Lines[angle=45,distance=6pt,line width=0.3pt]}, pattern color=black}]{2-1}{} \\
									&  &  \\
		\hline
		\Block[fill=blue!32, tikz={pattern={Lines[angle=45,distance=6pt,line width=0.3pt]}, pattern color=black}]{1-2}{} &  &  
	\end{bNiceMatrix}
	}_{\displaystyle \dbP(0)}
\end{align}
reveals the \emph{asymmetric} behaviour of~\eqref{eq:dP}: i) state uncertainty drives the state auto-covariance dynamics (top-left, red); ii) parameter uncertainty propagates into the cross-covariance blocks (off-diagonal, blue); iii) the parameter auto-covariance (bottom-right) has zero dynamics. Therefore, parameter uncertainty influences state evolution, but not vice versa.

The analytical solution of~\eqref{eq:dP}, verifiable by differentiation~\cite{Gajic:LyapunovMatrixEquation:1995}, takes the form
\begin{align}\label{eq:P_from_STM}
\bP(t) = \STM(t,t_0) \bP_0 \STM^T(t,t_0)+\int_{t_0}^{t} \STM(t,\tau) \bQ(\tau) \STM^T(t,\tau) \dd \tau \quad \bP(t_0)=\bP_0,
\end{align}
where $\STM(t,t_0)$ is the \emph{state transition matrix} (STM). This matrix encodes the evolution from $t_0$ to $t$ of a perturbation with respect to the mean trajectory $\bmu(t)$ as $\bar{\aug}(t) = \STM(t,t_0) \bar{\aug}(t_0)$, with $\bar{\aug}(\cdot) =\aug(\cdot) - \bmu(\cdot) $. 
The evolution of $\STM(t,t_k)$ from a generic $t_k$ to $t$ is driven by the following differential equation
\begin{align}\label{eq:dSTM}
\dSTM(t,t_k) = \bA(t)\STM(t,t_k), \quad \STM(t_k, t_k) = \bI.
\end{align}
Rather than integrating~\eqref{eq:dP} directly, we propagate the STM via~\eqref{eq:dSTM} and recover $\bP(t)$ a posteriori through~\eqref{eq:P_from_STM}, a well-established approach in geometric and orbital mechanics for statistical trajectory determination~\cite{Maruskin:DynamicalSystemsGeometric:2018,Tapley:StatisticalOrbitDetermination:2004}. The advantage of this choice in our framework will become apparent in Section~\ref{subsec:multiple_horizon_prop}.

\subsection{Robust friction limit constraint formulation} \label{subsec:friction_limit}

The model described in Section~\ref{subsec:vehicle_model} employs a pure lateral Magic Formula, while longitudinal forces are treated as control inputs. 
Since this formulation decouples longitudinal and lateral force generation, it does not guarantee that the resulting ground force is physically feasible when both components are required simultaneously. 
To account for combined-slip behaviour, we enforce that the total ground-reaction forces at each axle remain within the tyre adherence ellipse.
 
This friction limit constraint is expressed as 
\begin{equation}
    \bar{h}_j(\aug,\bu) =  S_j(\aug,\bu) - 1 \leq 0, \qquad{(j=1,2)}
\label{eq:friction_limit}
\end{equation}
where $S_j(\aug,\bu)$ is the \emph{axle saturation ratio}
\begin{equation}
    S_j(\aug,\bu) = \frac{ \left( \frac{X_j(\aug,\bu)}{\mu_{x,j}} \right)^2+ \left( \frac{Y_j(\aug,\bu)}{\mu_{y,j}}\right)^2}{Z_j^2(\aug,\bu)}.
\label{eq:axle_saturation}
\end{equation}
In Equations~\eqref{eq:friction_limit}--\eqref{eq:axle_saturation}, $j=1,2$ refers to the front and rear axles, respectively; $X_j$ and $Y_j$ denote the longitudinal and lateral components of the in-plane ground forces (Figure~\ref{fig:single_track}); $Z_j$ is the total vertical load on the axle; $\mu_{x,j}$ and $\mu_{y,j}$ are the longitudinal and lateral friction coefficients, both set to $1.15$ for all axles.
The \emph{axle saturation ratio} is equal to 0 when the in-plane ground force demand on the axle is zero, while it is equal to 1 when the point $\left(X_j, Y_j\right)$ lies exactly on the friction ellipse, i.e., under full saturation.
 
Since $\aug$ is a random variable, $\bar{h}_j(\aug,\bu)$ is itself random.
We therefore impose the chance constraint
$\Prob\big\{\bar{h}_j(\aug,\bu) \leq 0\big\} \geq p$,
where $p$ is a prescribed confidence level. We reformulate this chance constraint as a
deterministic constraint on the mean $\bmu$, at the cost of introducing a
back-off term that tightens the nominal constraint.
The back-off is defined as $\be_j = \ga \sig_j$, where $\ga=\Phi^{-1}(p)$ is the standard-normal quantile corresponding to the required probability $p$ of satisfying the constraint, and $\sig_j(\bmu, \bu, \bP) = \big[\na_{\bmu} \bar{h}_j(\bmu)^T \, \bP \, \na_{\bmu} \bar{h}_j(\bmu)\big]^{\frac{1}{2}}$ is the standard deviation of $\bar{h}_j$ linearised around the mean $\bmu$. The robust constraint is
\begin{equation}
    h_j(\bmu,\bu) = \bar{h}_j(\bmu,\bu) + \be_j(\bmu, \bP, \bu) \leq 0, \qquad j=1,2.
\label{eq:friction_limit_backoff}
\end{equation}

%% file: Fig/TikZ_Sources/tikz_vehicle_topview.tex
\tikzset{every picture/.style={line width=0.75pt}} 

\begin{tikzpicture}[x=0.75pt,y=0.75pt,yscale=-1,xscale=1]

\draw  [fill={rgb, 255:red, 148; green, 140; blue, 140 }  ,fill opacity=1 ][line width=0.75]  (349.48,127.72) .. controls (351.34,126.16) and (354.11,126.4) .. (355.67,128.26) -- (364.16,138.37) .. controls (365.72,140.23) and (365.48,143.01) .. (363.61,144.57) -- (331.47,171.54) .. controls (329.61,173.1) and (326.83,172.86) .. (325.27,171) -- (316.79,160.89) .. controls (315.23,159.03) and (315.47,156.25) .. (317.33,154.69) -- cycle ;
\draw    (318.4,159.64) -- (352.24,131.09) ;
\draw    (322.48,164.48) -- (340.27,149.47) -- (356.32,135.92) ;
\draw    (326.56,169.31) -- (345.32,153.48) -- (360.4,140.76) ;

\draw  [fill={rgb, 255:red, 148; green, 140; blue, 140 }  ,fill opacity=1 ][line width=0.75]  (191.09,189.08) .. controls (193.23,188.39) and (195.53,189.56) .. (196.22,191.71) -- (199.99,203.35) .. controls (200.68,205.49) and (199.51,207.79) .. (197.36,208.48) -- (152.75,222.92) .. controls (150.6,223.61) and (148.31,222.44) .. (147.61,220.29) -- (143.85,208.65) .. controls (143.15,206.51) and (144.33,204.21) .. (146.47,203.52) -- cycle ;
\draw    (148.86,206.46) -- (190.86,192.46) ;
\draw    (150.86,212.46) -- (172.94,205.1) -- (192.86,198.46) ;
\draw    (152.86,218.46) -- (176.14,210.7) -- (194.86,204.46) ;

\draw  [fill={rgb, 255:red, 185; green, 228; blue, 249 }  ,fill opacity=1 ] (457,149.6) -- (472,194.6) -- (484,240.6) -- (438,251.6) -- (437,248.6) -- (445,245.6) -- (434,205.6) -- (422,168.6) -- (406,125.6) -- (398,128.6) -- (397,124.6) -- (440,107.6) -- cycle ;
\draw  [fill={rgb, 255:red, 182; green, 218; blue, 236 }  ,fill opacity=1 ] (340,278.6) -- (339,285.6) -- (249,314.6) -- (207,187.6) -- (296,157.6) -- (301,163.6) -- (309,164.6) -- (345,273.6) -- cycle ;
\draw  [fill={rgb, 255:red, 198; green, 231; blue, 248 }  ,fill opacity=1 ] (380,175.8) .. controls (361.5,178.8) and (319,188.8) .. (283,197.8) .. controls (247,206.8) and (232,211.8) .. (203,222.8) .. controls (174,233.8) and (194,293.8) .. (224,288.2) .. controls (254,282.6) and (259,277.6) .. (299,263.6) .. controls (339,249.6) and (371,231) .. (388,224) .. controls (405,217) and (446,199) .. (450,193) .. controls (454,187) and (447,168.2) .. (441,165.6) .. controls (435,163) and (398.5,172.8) .. (380,175.8) -- cycle ;
\draw  [fill={rgb, 255:red, 185; green, 226; blue, 247 }  ,fill opacity=1 ] (124,234.2) -- (196.36,210.48) -- (227.09,302.08) -- (154,326) -- (153,323.2) -- (161,320.2) -- (133,234.2) -- (124.76,236.59) -- cycle ;
\draw [color={rgb, 255:red, 155; green, 155; blue, 155 }  ,draw opacity=0.36 ][line width=0.5]    (191,263.4) -- (154.14,149.39) ;
\draw [color={rgb, 255:red, 155; green, 155; blue, 155 }  ,draw opacity=0.36 ][line width=0.5]    (273.75,235.74) -- (237,124) ;
\draw [color={rgb, 255:red, 155; green, 155; blue, 155 }  ,draw opacity=0.36 ][line width=0.5]    (344.5,152.35) -- (327,98) ;
\draw  [fill={rgb, 255:red, 148; green, 140; blue, 140 }  ,fill opacity=1 ][line width=0.75]  (394.16,238.86) .. controls (396.03,237.3) and (398.8,237.54) .. (400.36,239.4) -- (408.85,249.51) .. controls (410.41,251.37) and (410.16,254.15) .. (408.3,255.71) -- (376.16,282.68) .. controls (374.3,284.24) and (371.52,284) .. (369.96,282.14) -- (361.48,272.03) .. controls (359.92,270.17) and (360.16,267.39) .. (362.02,265.83) -- cycle ;
\draw [fill={rgb, 255:red, 148; green, 140; blue, 140 }  ,fill opacity=1 ][line width=1.5]    (206.7,310.4) -- (175.3,216.4) ;
\draw  [fill={rgb, 255:red, 171; green, 206; blue, 223 }  ,fill opacity=1 ]  (260,214.6) .. controls (294,203.6) and (310,197.6) .. (317,221.6) .. controls (324,245.6) and (315,247.6) .. (276,260.6) .. controls (237,273.6) and (236,275.6) .. (228,251.1) .. controls (220,226.6) and (226,225.6) .. (260,214.6) -- cycle ;
\draw [fill={rgb, 255:red, 148; green, 140; blue, 140 }  ,fill opacity=1 ][line width=1.5]    (363,205.9) -- (191,263.4) ;
\draw [color={rgb, 255:red, 208; green, 2; blue, 27 }  ,draw opacity=1 ][line width=1.5]    (273.75,235.74) -- (312.11,222.92) ;
\draw [shift={(317,221.3)}, rotate = 161.55] [fill={rgb, 255:red, 208; green, 2; blue, 27 }  ,fill opacity=1 ][line width=0.08]  [draw opacity=0] (10,-4.8) -- (0,0) -- (10,4.8) -- cycle    ;
\draw  [draw opacity=0][line width=1.5]  (295.92,231.29) .. controls (296.35,233.8) and (296.35,236.44) .. (295.86,239.1) .. controls (293.61,251.31) and (281.88,259.38) .. (269.67,257.13) .. controls (257.46,254.88) and (249.38,243.16) .. (251.63,230.95) .. controls (252.12,228.29) and (253.06,225.83) .. (254.36,223.63) -- (273.75,235.74) -- cycle ; \draw [color={rgb, 255:red, 128; green, 0; blue, 200 }  ,draw opacity=1 ][line width=1.5]    (296.12,237.27) .. controls (296.06,237.87) and (295.97,238.48) .. (295.86,239.1) .. controls (293.61,251.31) and (281.88,259.38) .. (269.67,257.13) .. controls (257.46,254.88) and (249.38,243.16) .. (251.63,230.95) .. controls (252.12,228.29) and (253.06,225.83) .. (254.36,223.63) ;  \draw [shift={(295.92,231.29)}, rotate = 95.8] [fill={rgb, 255:red, 128; green, 0; blue, 200 }  ,fill opacity=1 ][line width=0.08]  [draw opacity=0] (10,-4.8) -- (0,0) -- (10,4.8) -- cycle    ;
\draw [color={rgb, 255:red, 23; green, 83; blue, 159 }  ,draw opacity=1 ][line width=2.25]    (191,263.4) -- (162.54,175.76) ;
\draw [shift={(161,171)}, rotate = 72.01] [fill={rgb, 255:red, 23; green, 83; blue, 159 }  ,fill opacity=1 ][line width=0.08]  [draw opacity=0] (14.29,-6.86) -- (0,0) -- (14.29,6.86) -- cycle    ;
\draw [color={rgb, 255:red, 200; green, 200; blue, 200 }  ,draw opacity=1 ][line width=0.5]    (244.79,135.72) -- (326.21,108.28) ;
\draw [color={rgb, 255:red, 200; green, 200; blue, 200 }  ,draw opacity=1 ][line width=0.5]    (161.94,163.15) -- (238.2,138.24) ;
\draw [color={rgb, 255:red, 74; green, 74; blue, 74 }  ,draw opacity=1 ][line width=0.75]    (421,205.9) -- (363,205.9) ;
\draw  [draw opacity=0] (412.71,190.67) .. controls (414.56,195.5) and (415.59,200.65) .. (415.7,205.9) -- (370.22,206.56) -- cycle ; \draw    (413.7,193.53) .. controls (414.93,197.51) and (415.61,201.67) .. (415.7,205.9) ;  \draw [shift={(412.71,190.67)}, rotate = 76.69] [fill={rgb, 255:red, 0; green, 0; blue, 0 }  ][line width=0.08]  [draw opacity=0] (10.72,-5.15) -- (0,0) -- (10.72,5.15) -- (7.12,0) -- cycle    ;
\draw [fill={rgb, 255:red, 148; green, 140; blue, 140 }  ,fill opacity=1 ][line width=1.5]    (379,252.6) -- (347,159.2) ;
\draw  [fill={rgb, 255:red, 148; green, 140; blue, 140 }  ,fill opacity=1 ][line width=0.75]  (229.09,303.08) .. controls (231.23,302.39) and (233.53,303.56) .. (234.22,305.71) -- (237.99,317.35) .. controls (238.68,319.49) and (237.51,321.79) .. (235.36,322.48) -- (190.75,336.92) .. controls (188.6,337.61) and (186.31,336.44) .. (185.61,334.29) -- (181.85,322.65) .. controls (181.15,320.51) and (182.33,318.21) .. (184.47,317.52) -- cycle ;
\draw    (113,364.2) -- (270.75,364.2) ;
\draw [shift={(273.75,364.2)}, rotate = 180] [fill={rgb, 255:red, 0; green, 0; blue, 0 }  ][line width=0.08]  [draw opacity=0] (8.93,-4.29) -- (0,0) -- (8.93,4.29) -- cycle    ;
\draw    (113,364.2) -- (113,238.74) ;
\draw [shift={(113,235.74)}, rotate = 90] [fill={rgb, 255:red, 0; green, 0; blue, 0 }  ][line width=0.08]  [draw opacity=0] (8.93,-4.29) -- (0,0) -- (8.93,4.29) -- cycle    ;
\draw [color={rgb, 255:red, 160; green, 160; blue, 160 }, draw opacity=0.7, line width=0.5]  (273.75,364.2) -- (273.75,235.74) ;
\draw [color={rgb, 255:red, 160; green, 160; blue, 160 }, draw opacity=0.7, line width=0.5]  (113,235.74) -- (273.75,235.74) ;
\draw [color={rgb, 255:red, 23; green, 83; blue, 159 }  ,draw opacity=1 ][line width=2.25]    (363,205.9) -- (334.6,121.74) ;
\draw [shift={(333,117)}, rotate = 71.35] [fill={rgb, 255:red, 23; green, 83; blue, 159 }  ,fill opacity=1 ][line width=0.08]  [draw opacity=0] (14.29,-6.86) -- (0,0) -- (14.29,6.86) -- cycle    ;
\draw [color={rgb, 255:red, 74; green, 74; blue, 74 }  ,draw opacity=1 ][line width=0.75]    (420,189) -- (363,205.9) ;
\draw [color={rgb, 255:red, 208; green, 2; blue, 27 }  ,draw opacity=1 ][line width=1.5]    (273.75,235.74) -- (265.3,209.6) ;
\draw [shift={(263.7,204.9)}, rotate = 71.55] [fill={rgb, 255:red, 208; green, 2; blue, 27 }  ,fill opacity=1 ][line width=0.08]  [draw opacity=0] (10,-4.8) -- (0,0) -- (10,4.8) -- cycle    ;
\draw [color={rgb, 255:red, 74; green, 74; blue, 74 }  ,draw opacity=1 ][line width=0.75]    (381.5,114.45) -- (343.47,146.54) ;
\draw [color={rgb, 255:red, 74; green, 74; blue, 74 }  ,draw opacity=1 ][line width=0.75]    (400,130) -- (343,146.9) ;
\draw  [draw opacity=0] (376.2,118.44) .. controls (380.02,123.11) and (382.87,128.48) .. (384.59,134.21) -- (341,146.9) -- cycle ; \draw    (378.06,120.88) .. controls (380.95,124.94) and (383.16,129.45) .. (384.59,134.21) ;  \draw [shift={(376.2,118.44)}, rotate = 58.3] [fill={rgb, 255:red, 0; green, 0; blue, 0 }  ][line width=0.08]  [draw opacity=0] (10.72,-5.15) -- (0,0) -- (10.72,5.15) -- (7.12,0) -- cycle    ;
\draw [color={rgb, 255:red, 23; green, 83; blue, 159 }  ,draw opacity=1 ][line width=2.25]    (191,263.4) -- (231.28,249.26) ;
\draw [shift={(236,247.6)}, rotate = 160.65] [fill={rgb, 255:red, 23; green, 83; blue, 159 }  ,fill opacity=1 ][line width=0.08]  [draw opacity=0] (14.29,-6.86) -- (0,0) -- (14.29,6.86) -- cycle    ;
\draw [color={rgb, 255:red, 23; green, 83; blue, 159 }  ,draw opacity=1 ][line width=2.25]    (363,205.9) -- (337.73,214.49) ;
\draw [shift={(333,216.1)}, rotate = 341.22] [fill={rgb, 255:red, 23; green, 83; blue, 159 }  ,fill opacity=1 ][line width=0.08]  [draw opacity=0] (14.29,-6.86) -- (0,0) -- (14.29,6.86) -- cycle    ;
\draw    (221,260.6) .. controls (212,288) and (257,305) .. (248,332.6) ;
\draw    (343,221.6) .. controls (336,250) and (370,271) .. (363,299.6) ;
\draw    (187,320.6) -- (229,306.6) ;
\draw    (189,326.6) -- (211.08,319.24) -- (231,312.6) ;
\draw    (191,332.6) -- (214.28,324.84) -- (233,318.6) ;

\draw    (363.4,270.64) -- (397.24,242.09) ;
\draw    (367.48,275.48) -- (385.27,260.47) -- (401.32,246.92) ;
\draw    (371.56,280.31) -- (390.32,264.48) -- (405.4,251.76) ;

\draw [line width=0.75]    (139,232.2) -- (167,318.2) ;
\draw [line width=0.75]    (147,229.2) -- (175,315.2) ;
\draw [line width=0.75]    (213.67,191.2) -- (221.33,215.27) ;
\draw [line width=0.75]    (220.67,189.2) -- (228.33,213.27) ;
\draw [line width=0.75]    (243.67,283.2) -- (251.33,307.27) ;
\draw [line width=0.75]    (250.67,281.2) -- (258.33,305.27) ;

\draw (294,209) node [anchor=north west][inner sep=0.75pt]    {$u$};
\draw (262.95,191) node [anchor=north west][inner sep=0.75pt]    {$v$};
\draw (242,249.42) node [anchor=north west][inner sep=0.75pt]    {$r$};
\draw (334,104) node [anchor=north west][inner sep=0.75pt]    {${\displaystyle Y_{1}}$};
\draw (143,160) node [anchor=north west][inner sep=0.75pt]    {${\displaystyle Y_{2}}$};
\draw (384.27,109.59) node [anchor=north west][inner sep=0.75pt]    {$\delta $};
\draw (270,107) node [anchor=north west][inner sep=0.75pt]  [rotate=-341.46]  {$a_{1}$};
\draw (185,136) node [anchor=north west][inner sep=0.75pt]  [rotate=-339.83]  {$a_{2}$};
\draw (422.15,184.96) node [anchor=north west][inner sep=0.75pt]    {$\psi $};
\draw (246.5,334.9) node [anchor=north west][inner sep=0.75pt]    {${\displaystyle X_{2}}$};
\draw (364.38,309.69) node    {${\displaystyle X_{1} < 0}$};
\draw (271,370) node [anchor=north west][inner sep=0.75pt]    {$x_G$};
\draw (78,228.72) node [anchor=north west][inner sep=0.75pt]    {$y_G$};

\draw  [fill={rgb, 255:red, 24; green, 23; blue, 23 }  ,fill opacity=1 ] (273.75,233.03) .. controls (275.24,233.03) and (276.46,234.24) .. (276.46,235.74) .. controls (276.46,237.24) and (275.24,238.45) .. (273.75,238.45) .. controls (272.25,238.45) and (271.03,237.24) .. (271.03,235.74) .. controls (271.03,234.24) and (272.25,233.03) .. (273.75,233.03) -- cycle ;
\draw (272.23,215.94) node [anchor=north west][inner sep=0.75pt]    {$G$};

\end{tikzpicture}

%% file: robust_mltp.tex
\section{Robust minimum-lap-time planning} \label{sec:robust_mltp}
We formulate the Optimal Control Problem (OCP) encoding
robust minimum-lap-time planning on a motorsport track.
We transcribe it into a discrete nonlinear programme (NLP) by applying the direct
collocation approach~\cite{Gillis:PracticalMethodsApproximate:2015}. 
The track is parameterised along its centreline by a curvilinear coordinate $\alpha \in [0,1]$ and sampled at $N+1$ grid nodes $\alpha_0, \ldots, \alpha_N$. Within each interval $[\alpha_{k-1},
\alpha_k]$, the state trajectories are approximated by polynomials defined at $d$
collocation points on the unit interval.

\subsection{Multiple short-horizon propagation} \label{subsec:multiple_horizon_prop}
Planning under state and parameter uncertainty requires propagating the covariance $\bP$. Since we tackle an open-loop planning problem, i.e., without feedback policies, position and orientation uncertainty accumulates along the track. 
Over a \emph{medium-to-long} spatial horizon, e.g., a circuit sector, the covariance grows without bound and produces an overly conservative back-off. 

Therefore, we propagate the covariance matrix for a \emph{short prediction horizon}, starting from each grid point $k$ of the discretised trajectory and accumulating uncertainty only over the following $H$ steps, with $H$ chosen small enough to capture the disturbance evolution before any corrective driver action, which the planning framework does not model.

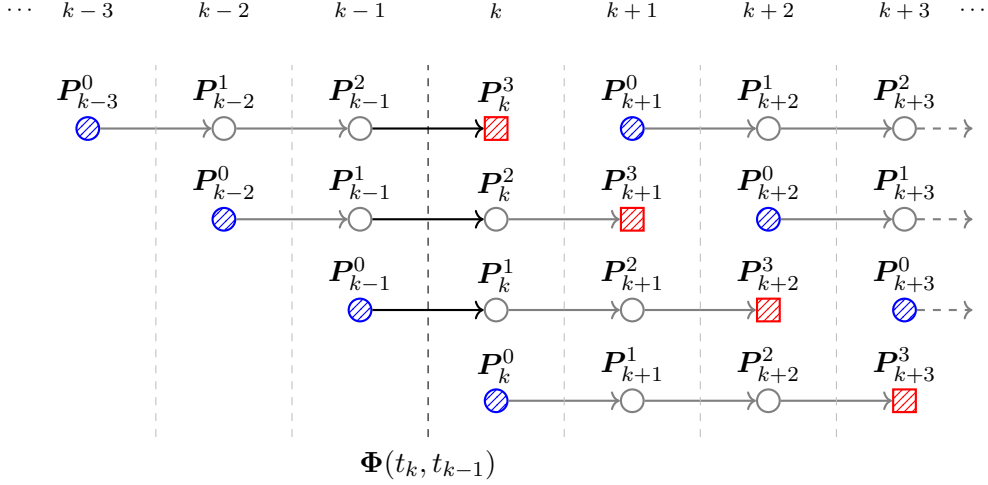
\begin{figure}
	\centering
	\definecolor{mygreen}{RGB}{0,150,0}
	\input{Fig/TikZ_Sources/tikz_covariance_evolution.tex}
	\caption{Multiple short-horizon propagation of the covariance matrix, shown here for $H=3$. Each node is a covariance matrix instance: the subscript fixes the grid index, whilst the superscript counts the steps elapsed since that instance was initialised. At every step, $H+1$ instances coexist.Along each propagation chain, the initialised covariance (blue node) and the end-of-horizon instance (red square)---on which we enforce robust constraints---are highlighted. Arrows crossing the same column boundary all carry the identical state-transition matrix, e.g., $\STM(t_k,t_{k-1})$ from $t_{k-1}$ to $t_k$.}
	\label{fig:multiple_hor_prop}
\end{figure}

Under this strategy, each grid point carries \emph{multiple versions} of the covariance matrix. To this end, we introduce the notation $\bP_k^j$, where $j = 0, \ldots, H$, and $\bP_k^j$ represents the version of the covariance matrix at step $k$ that was initialised $j$ steps earlier. 
Figure~\ref{fig:multiple_hor_prop} illustrates the multiple instances of the covariance matrix.
The dashed rectangle highlights the $k$-th discretisation step.
At each grid point, two particular versions of the covariance matrix are emphasised: $\bP^0_k$ (red node), representing the instance to be initialised at that step, and $\bP^H_k$ (green node), corresponding to the version that has been propagated over the full horizon of $H$ steps. This end-of-horizon instance $\bP^H_k$ is precisely the covariance matrix used to tighten at step $k$ the friction limit constraint introduced in Section~\ref{subsec:friction_limit}.

\subsection{OCP formulation} \label{subsec:ocp}
We introduce the following notation for the decision variables in the OCP.
At each grid node $k$, $\bmu_k$ denotes the mean augmented state and $\STM_k=\STM(t_k,t_{k-1})$ the state
transition matrix; the corresponding values at the $d$ collocation points within
interval $k$ are collected in $\bxi_k = (\bxi_{k,1}, \ldots, \bxi_{k,d})$ and
$\bSi_k = (\bSi_{k,1}, \ldots, \bSi_{k,d})$, respectively.
Here, $\bSi_{k,i}=\STM(t_{k,i},t_{k-1})$ is the STM at the $i$-th collocation point within the $k$-th interval $[t_{k-1},t_k]$.
The inputs $\bu_k$ and algebraic variables $\bz_k$---including ground reaction forces and the covariance matrices---are defined at the grid nodes only and held piecewise constant over each interval.

The resulting OCP has the following formulation:
\begin{subequations} \label{eq:ocp}
	\begin{alignat}{3}
		\hspace*{-2.0 cm}\llap{$\displaystyle\underset{\substack{\bmu_k,\bxi_k, \bu_k, \\ \STM_k, \bSi_k, \bz_k}}{\text{minimise}}$}\,
		& & & J_k(\bmu_k,\bxi_k, \bu_k) & & \label{eq:ocp:cost} \\
		\hspace*{-2.0 cm}\text{s.t.} \quad
		& \bzero      & = & \; \bPsimu_k(\bmu_{k-1},\bmu_k,\bxi_k, \bu_k,\bz_k),
		& \quad & k \in [1,N] \label{eq:ocp:dyn} \\
		& \bmu_0      & = & \; \bar{\bmu}_0
		& & \label{eq:ocp:dyn_ic} \\
		& \bzero      & = & \; \bPsiSTM_k(\bxi_k,\STM_k,\bSi_k,\bu_k,\bz_k),
		& \quad & k \in [1,N] \label{eq:ocp:stm} \\
		& \bzero      & = & \; \bOm_k(\bmu_k,\bxi_k,\STM_k,\bSi_k, \bu_k,\bz_k),
		& \quad & k \in [0,N] \label{eq:ocp:alg} \\
		& 0           & \geq & \; h_i(\bmu_k, \bu_k, \bz_k) + \be_i(\bmu_k, \bu_k, \bz_k),
		& \quad & k \in [1,N],\; i \in \calI \quad (\lambda_{k,i} \geq 0) \label{eq:ocp:constr}
	\end{alignat}
\end{subequations}
The cost~\eqref{eq:ocp:cost} encodes the minimum-lap-time objective.
Equation~\eqref{eq:ocp:dyn} includes the collocation and continuity conditions for the mean dynamics~\eqref{eq:mean_dyn},
initialised at $\bar{\bmu}_0$ in~\eqref{eq:ocp:dyn_ic}.
Equation~\eqref{eq:ocp:stm} 
encodes the collocation and continuity conditions for the dynamics of the STM~\eqref{eq:dSTM}.
Equation~\eqref{eq:ocp:alg} collects the path equality constraints,
which include the algebraic recovery of the covariance instances $\bP_k^j$ from
$\STM_k$ via~\eqref{eq:P_from_STM}.
Finally, \eqref{eq:ocp:constr} enforces the robust inequality constraints, with
$\calI$ the set of indices of the inequality constraints, and $\lambda_{k,i}$ the Lagrange multiplier associated with the $i$-th constraint at the $k$-th grid node. 
In our case, \eqref{eq:ocp:constr} includes the robust friction limit constraints defined in~\eqref{eq:friction_limit_backoff}. 
The back-off $\be_i$ is evaluated using the end-of-horizon covariance $\bP_k^H$, whose elements are included in $\bz_k$.

To recover the nominal---non-stochastic---MLTP formulation, it suffices to remove the STM dynamics from~\eqref{eq:ocp} and the back-off terms from~\eqref{eq:ocp:constr}. 
With respect to the nominal MLTP problem, the robust formulation in~\eqref{eq:ocp} introduces the STMs as additional decision variables at all grid and collocation nodes---each STM instance adding $n_x(n_x+n_p)$ scalar variables, since the lower $n_p$ rows of each STM are structurally constant and equal to $[\bzero \mid \bI_{n_p}]$, by an argument analogous to~\eqref{eq:dP_color1}--\eqref{eq:dP_color2}.
Thus, the robust OCP incurs a considerably higher computational cost than the nominal MLTP formulation.

\subsection{Parsimonious formulation} \label{subsec:parsimonious}
Starting from the nominal MLTP solution, we build a \emph{parsimonious} variant of~\eqref{eq:ocp} that activates the robust machinery (state-transition matrices, covariance propagation, and constraint back-off) only at nodes where the nominal trajectory operates near the friction limit.
Elsewhere, the back-off would tighten constraints already satisfied with ample margin, yielding no safety benefit.
Confining robustness to the nodes at or near saturation therefore reduces the number of decision variables in the OCP, avoiding unnecessary STM and covariance instances and thereby lowering the computational cost relative to the full robust formulation.

Let $\calC=\{1,\dots,N\}$ be the set of grid nodes. Among the inequality constraints $i\in\calI$, we focus on the friction-limit constraints~\eqref{eq:friction_limit}, indexed by axle $j\in\{1,2\}$ (front and rear). At each node $k$, we read from the nominal solution two indicators of these constraints.

The first is the nominal Lagrange multiplier~\cite{Nocedal:NumericalOptimization:2006} $\lambda_{k,j}^{(\mathrm{nom})}\ge 0$, which is positive only on the \emph{active set}, i.e.,\ where the axle lies exactly on the friction boundary. At the NLP optimum, its magnitude is the \emph{shadow price} of that constraint: the larger $\lambda_{k,j}^{(\mathrm{nom})}$, the larger the lap-time reduction attainable by relaxing the corresponding friction limit; it therefore also ranks the saturated nodes according to how strongly they constrain performance. 

The second is the constraint residual $\barh_{k,j}=\bar{h}_j(\bmu_k,\bu_k,\bz_k)$, the function~\eqref{eq:friction_limit} evaluated at node $k$ (hence the added subscript $k$): zero at saturation and negative inside the friction ellipse---the more negative, the larger the margin from the friction limit. We summarise each node by its worst axle,
\begin{equation}\label{eq:node_residual}
	\barh_k = \max_{j\in\{1,2\}} \barh_{k,j} ,
\end{equation}
so that $\barh_k$ retains the larger residual of the front ($j=1$) and rear ($j=2$) axles, yielding a single residual per node.

We build the parsimonious set $\calCpar$ in two steps. First, the \emph{critical} (saturated) nodes are those carrying an active friction constraint,
\begin{equation}\label{eq:Csat}
	\calCsat = \bigl\{\, k \in \calC : \exists\, j\in\{1,2\},\; \lambda_{k,j}^{(\mathrm{nom})} > 0 \,\bigr\} ,
\end{equation}
where, in practice, ``positive'' means above a small numerical tolerance, since interior-point solvers return slightly positive multipliers. Having no residual margin, these nodes are robustified unconditionally.

Critical nodes alone, however, leave unprotected the nodes that operate near the friction limit without reaching it, where a disturbance could push the axle to saturation. We therefore introduce a small additional \emph{budget} of near-critical nodes, sized as a fraction $\rho\in(0,1)$ of the grid, $n_\rho=\mathrm{ceil}(\rho N)$, where $\mathrm{ceil}(\cdot)$ rounds up to the next integer so that the budget comprises at least one node; for instance, $\rho=0.05$ (a $5\%$ budget) on a grid of $N=100$ nodes yields $n_\rho=5$ nodes. This budget is allocated to the non-critical nodes closest to the limit, that is, the $n_\rho$ nodes of $\calC\setminus\calCsat$ with the largest (less negative) residual,
\begin{equation}\label{eq:Cthr}
	\calCthr = \bigl\{\, k \in \calC\setminus\calCsat : \barh_k \ge \barh_{\mathrm{thr}} \,\bigr\} ,
\end{equation}
where $\barh_{\mathrm{thr}}$ is the $n_\rho$-th largest residual among those nodes. The fraction $\rho$ thus acts as a robustness budget, setting how many nodes, beyond the strictly saturated ones, are treated as robust.

The parsimonious OCP keeps the full structure of~\eqref{eq:ocp} but carries the STM and covariance variables, and the back-off, only on
\begin{equation}\label{eq:Cpar}
  \calCpar = \calCsat \cup \calCthr \subseteq \calC ,
\end{equation}
reducing the number of decision variables relative to the full robust formulation, where they appear at every node. The result is a computationally lighter OCP that retains the solution quality of the full robust formulation, as Section~\ref{sec:experiments_setup} will detail.

%% file: Fig/TikZ_Sources/tikz_covariance_evolution.tex
\begin{tikzpicture}[
	smallnode/.style={circle, draw=gray, minimum size=3mm, inner sep=0pt},
	r_smallnode/.style={circle, draw=blue, minimum size=3mm, inner sep=0pt},
	g_smallnode/.style={circle, draw=red, minimum size=3mm, inner sep=0pt},
	r_hatnode/.style={
		circle, draw=blue, pattern=north east lines, pattern color=blue,
		minimum size=3mm, inner sep=0pt},
	g_hatnode/.style={
		rectangle, draw=red, pattern=north east lines, pattern color=red,
		minimum size=3mm, inner sep=0pt},
	every path/.style={->, thick},
	x=1.8cm, y=1.2cm
	]
	
	\node[r_hatnode, label={[label distance=-2pt]above:\(\bP_{k-3}^0\)}] (P00) at (1,3) {};
	\node[smallnode, label={[label distance=-2pt]above:\(\bP_{k-2}^1\)}] (P11) at (2,3) {};
	\node[smallnode, label={[label distance=-2pt]above:\(\bP_{k-1}^2\)}] (P22) at (3,3) {};
	\node[g_hatnode, label={[label distance=-2pt]above:\(\bP_k^3\)}] (P33) at (4,3) {};
	\node[r_hatnode, label={[label distance=-2pt]above:\(\bP_{k+1}^0\)}] (P44) at (5,3) {};
	\node[smallnode, label={[label distance=-2pt]above:\(\bP_{k+2}^1\)}] (P55) at (6,3) {};
	\node[smallnode, label={[label distance=-2pt]above:\(\bP_{k+3}^2\)}] (P66) at (7,3) {};
	
	\node[r_hatnode, label={[label distance=-2pt]above:\(\bP_{k-2}^0\)}] (P10) at (2,2) {};
	\node[smallnode, label={[label distance=-2pt]above:\(\bP_{k-1}^1\)}] (P21) at (3,2) {};
	\node[smallnode, label={[label distance=-2pt]above:\(\bP_k^2\)}] (P32) at (4,2) {};
	\node[g_hatnode, label={[label distance=-2pt]above:\(\bP_{k+1}^3\)}] (P43) at (5,2) {};
	\node[r_hatnode, label={[label distance=-2pt]above:\(\bP_{k+2}^0\)}] (P54) at (6,2) {};
	\node[smallnode, label={[label distance=-2pt]above:\(\bP_{k+3}^1\)}] (P65) at (7,2) {};

	\node[r_hatnode, label={[label distance=-2pt]above:\(\bP_{k-1}^0\)}] (P20) at (3,1) {};
	\node[smallnode, label={[label distance=-2pt]above:\(\bP_k^1\)}] (P31) at (4,1) {};
	\node[smallnode, label={[label distance=-2pt]above:\(\bP_{k+1}^2\)}] (P42) at (5,1) {};
	\node[g_hatnode, label={[label distance=-2pt]above:\(\bP_{k+2}^3\)}] (P53) at (6,1) {};
	\node[r_hatnode, label={[label distance=-2pt]above:\(\bP_{k+3}^0\)}] (P64) at (7,1) {};

	\node[r_hatnode, label={[label distance=-2pt]above:\(\bP_k^0\)}] (P30) at (4,0) {};
	\node[smallnode, label={[label distance=-2pt]above:\(\bP_{k+1}^1\)}] (P41) at (5,0) {};
	\node[smallnode, label={[label distance=-2pt]above:\(\bP_{k+2}^2\)}] (P52) at (6,0) {};
	\node[g_hatnode, label={[label distance=-2pt]above:\(\bP_{k+3}^3\)}] (P63) at (7,0) {};

	\begin{scope}[gray]
		\draw (P00) -- (P11);
		\draw (P11) -- (P22);

		\draw (P44) -- (P55);
		\draw (P55) -- (P66);
		\draw[dashed,->] (P66) -- ++(0.5,0);

		\draw (P10) -- (P21);
		\draw (P32) -- (P43);
		\draw (P54) -- (P65);
		\draw[dashed,->] (P65) -- ++(0.5,0);

		\draw (P31) -- (P42);
		\draw (P42) -- (P53);
		\draw[dashed,->] (P64) -- ++(0.5,0);

		\draw (P30) -- (P41);
		\draw (P41) -- (P52);
		\draw (P52) -- (P63);
	\end{scope}
	\draw[black] (P22) -- (P33);
	\draw[black] (P21) -- (P32);
	\draw[black] (P20) -- (P31);
	
	\draw[-, gray!50, thin, dashed] (1.5,-0.4) -- (1.5,3.7);
	\draw[-, gray!50, thin, dashed] (2.5,-0.4) -- (2.5,3.7);
	\draw[-, black, thin, dashed] (3.5,-0.4) -- (3.5,3.7);
	\draw[-, gray!50, thin, dashed] (4.5,-0.4) -- (4.5,3.7);
	\draw[-, gray!50, thin, dashed] (5.5,-0.4) -- (5.5,3.7);
	\draw[-, gray!50, thin, dashed] (6.5,-0.4) -- (6.5,3.7);

	\node at (0.5, 4.3) {\scriptsize \(\cdots\)};
	\node at (1, 4.3) {\scriptsize \(k-3\)};
	\node at (2, 4.3) {\scriptsize \(k-2\)};
	\node at (3, 4.3) {\scriptsize \(k-1\)};
	\node at (4, 4.3) {\scriptsize \(k\)};
	\node at (5, 4.3) {\scriptsize \(k+1\)};
	\node at (6, 4.3) {\scriptsize \(k+2\)};
	\node at (7, 4.3) {\scriptsize \(k+3\)};
	\node at (7.5, 4.3) {\scriptsize \(\cdots\)};

	\node at (3.5,-0.7) {\(\STM(t_k,t_{k-1})\)};
	
	%
	
\end{tikzpicture}

%% file: experiments_setup.tex
\section{Experiments setup} \label{sec:experiments_setup}
We validate the proposed framework through a simulation campaign. A model predictive controller (MPC) serves as a virtual driver, tracking MLTP references with different robustness settings over multiple closed-loop runs on a simulated FSAE vehicle; each run includes impulsive disturbances and parameter variations to probe reference driveability. Section~\ref{subsec:reference_planning} details the reference planning, Section~\ref{subsec:mpc_design} the MPC design, and Section~\ref{subsec:testing_strategy} the testing strategy.

\subsection{Reference trajectory planning} \label{subsec:reference_planning}
\begin{figure}
	\centering
	\includegraphics{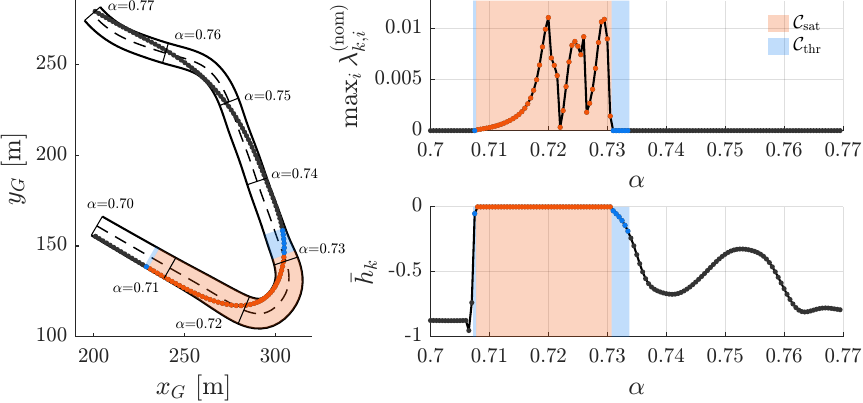}
	\caption{Left: NOM optimal trajectory on the analysed Catalunya sector ($\alpha\in[0.70,0.77]$), with the critical ($\calCsat$, orange) and near-critical ($\calCthr$, blue) node sets highlighted; checkpoints every $\Delta\alpha=0.01$. Top right: maximum Lagrange multiplier of the friction-limit constraints vs $\alpha$; $\calCsat$ (orange) collects the nodes where $\max_j\lambda_{k,j}^{(\text{nom})} > 0$. Bottom right: maximum constraint residual $\barh_k$ vs $\alpha$; $\calCthr$ (blue) collects the $n_\rho$ non-saturated nodes with residual closest to zero.}
	\label{fig:parsimonious}
\end{figure}

We apply the robust MLTP framework introduced in Section~\ref{sec:robust_mltp} to a FSAE vehicle driving on a representative sector of the Catalunya circuit (Figure~\ref{fig:parsimonious}).
The entire circuit is parameterised by the curvilinear parameter $\alpha \in [0,1]$, and the selected sector corresponds to the interval $\left[0.70, 0.77\right]$. 
This sector includes two distinct corners: a low-speed turn for $\alpha\in[0.72,0.73]$, and a high-speed turn for $\alpha\in[0.75,0.76]$.
We discretise the track section into $N=140$ spatial intervals---corresponding to a spatial resolution $\Delta s \approx 0.43$\,m---and build a grid of $N+1$ nodes for the OCP introduced in Problem~\eqref{eq:ocp}. The direct-collocation approach employs Gauss-Legendre collocation points and $d=2$ as collocation degree.

We consider four MLTP references. The nominal reference (NOM), which serves as the baseline, solves the deterministic MLTP with no uncertainty on states or parameters; it is recoverable from~\eqref{eq:ocp} by removing the STM dynamics and back-off terms.
ROB-S is robust with respect to the vehicle states only: the multiple short-horizon propagation scheme (Section~\ref{subsec:multiple_horizon_prop}) runs over all the grid nodes $k\in[1,N]$, and the robust inequality constraints, i.e., with back-off terms, hold on the whole track sector.
PAR-S retains state-only uncertainty but exploits the parsimonious formulation of Section~\ref{subsec:parsimonious}, enforcing the robust constraints only on the parsimonious set $\calCpar$ introduced in~\eqref{eq:Cpar}; hence, for each node $k\in\calCpar$, only the STM instances over the $H$ preceding steps and the resulting end-of-horizon covariance $\bP_k^H$ enter the OCP as decision variables.
PAR-SP extends PAR-S to joint state-and-parameter uncertainty.

In ROB-S, PAR-S and PAR-SP we adopt $H=5$ steps for the multiple short-horizon propagation---approximately $2$\,m of track, short enough to capture the disturbance evolution before a corrective driver response---and a confidence level $p=0.90$ (i.e., $\gamma=1.28$), a moderately conservative design choice for the back-off terms. 
Each one of the short-horizons of Figure~\ref{fig:multiple_hor_prop} is initialised at $\bP_k^0=\bP_0$, which controls the uncertainty on states and parameters together with the covariance matrix $\bQ$ of the additive noise $\bw$ defined in~\eqref{eq:dynamics}.
Assuming independent initial uncertainties on states and parameters, we impose a diagonal $\bP_0 = \diag(\bar{\bsig}_{\eta})^2$, with $\bar{\bsig}_{\eta}$ the standard deviations of the augmented state $\aug$ reported in Table~\ref{tab:P0}; for comparison, the nominal (mean) parameter values are listed in Table~\ref{tab:vehicle_params} (Section~\ref{subsec:vehicle_model}).
The additive noise $\bw$ models small, random disturbances to the vehicle dynamics---grip micro-variations, unmodelled aerodynamic effects---acting directly on the time derivatives of the velocity states $(u,v,r)$. We consequently set $\bQ = \diag([\bar{\bsig}_w,\,\bzero])^2$ with $\bar{\bsig}_w = [0.055,\;0.032,\;0.05]$ (m/s$^2$, m/s$^2$, rad/s$^2$) and zero entries for all remaining states and parameters. 
\begin{table}
	\centering\small
	\setlength{\tabcolsep}{4pt}
	\caption{Standard deviations $\bar{\bsig}_{\eta}$ of the initial covariance matrix $\bP_0 = \diag(\bar{\bsig}_{\eta})^2$ for robust references.}
	\label{tab:P0}
	\begin{tabular}{lcccccc|cccc}
		\toprule
		 & $u$ & $v$ & $r$ & $x_G$ & $y_G$ & $\psi$ & $J_z$ & $h$ & $w_b$ & $C_x$ \\
		 & {(m/s)} & {(m/s)} & {(rad/s)} & {(m)} & {(m)} & {(deg)} & {(kg\,m$^2$)} & {(m)} & {(--)} & {(--)} \\
		\midrule
		ROB-S, PAR-S & 0.20 & 0.06 & 0.05 & 0.50 & 0.50 & 1.0 & 0 & 0 & 0 & 0 \\
		PAR-SP       & 0.20 & 0.06 & 0.05 & 0.50 & 0.50 & 1.0 & 6.0 & 0.02 & 0.02 & 0.04 \\
		\bottomrule
	\end{tabular}
\end{table}

We transcribe the four OCPs in the CasADi--MATLAB environment~\cite{Andersson:CasADiSoftwareFramework:2019}, and solve the resulting nonlinear programs with IPOPT's interior-point algorithm~\cite{Wachter:ImplementationInteriorpointFilter:2006}. 
The nominal solution serves as warm start for the other three MLTPs and allows us to derive the critical and near-critical node sets $\calCsat$ and $\calCthr$ introduced in Section~\ref{subsec:parsimonious} for the parsimonious formulations used in PAR-S and PAR-SP. Figure~\ref{fig:parsimonious} shows the critical nodes $\calCsat$ (orange) and the near-critical nodes $\calCthr$ (blue) computed from NOM, with the optimal trajectory (left panel), maximum Lagrange multiplier of friction limit constraints (top right panel), and maximum constraint residual (bottom right panel) plotted as a function of the curvilinear abscissa $\alpha$.
Before solving each robust OCP, we run an intermediate \emph{warm-start solve}: starting from NOM, we augment the decision-variable set with the STM instances of~\eqref{eq:ocp} but leave the robust back-off terms inactive, so its solution both initialises the STM-related variables and supplies the converged starting point for the final, fully-constrained solve of ROB-S, PAR-S and PAR-SP. Because this solve only needs to seed the final problem, we run it at looser IPOPT tolerances, which is why it converges faster than NOM in Table~\ref{tab:solver_stats} despite its larger problem size.

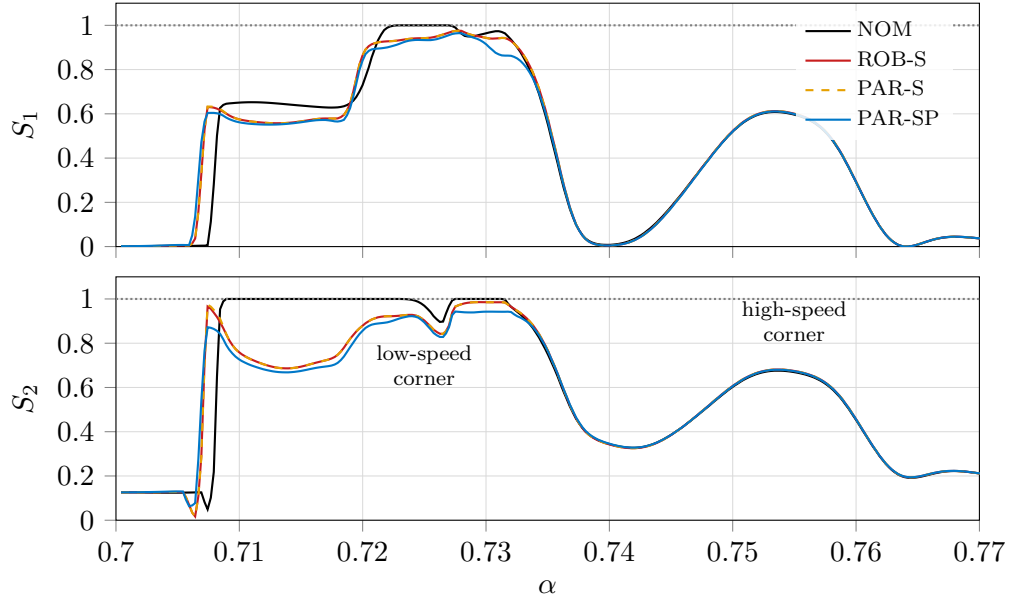
\begin{figure}
	\centering
	\input{Fig/TikZ_Sources/fig_ref_adherence}
	\caption{Front ($S_1$, top) and rear ($S_2$, bottom) axle saturation of the four MLTP reference solutions vs.\ track abscissa $\alpha$: NOM (solid black), ROB-S (solid red), PAR-S (dashed yellow), PAR-SP (solid blue). The dotted line marks the friction limit ($S_i=1$). PAR-S is shown dashed as it overlaps ROB-S.}
	\label{fig:adherence_ref}
  \end{figure}

Figure~\ref{fig:adherence_ref} shows the front and rear axle saturation indices---$S_1$ and $S_2$, respectively---along the track curvilinear abscissa $\alpha$, corresponding to the solution of NOM (solid black lines), ROB-S (solid red lines), PAR-S (dashed yellow lines), and PAR-SP (solid blue lines).
We report in Table~\ref{tab:solver_stats} the problem size, sparsity, IPOPT iterations and wall time for each MLTP, broken down into the warm-start solve and the final solve described above. The references are generated on a laptop with an Intel Core i7-12700H CPU at 2.30\,GHz.

\begin{table}
	\centering\small
	\setlength{\tabcolsep}{4pt}
	\caption{Problem size, IPOPT iterations and wall time for the four MLTP references, broken down into the warm-start solve (Section~\ref{subsec:reference_planning}) and the final, fully-constrained solve. ``Nodes with robust constraint'' is the fraction of grid nodes carrying the robust back-off constraints, reduced under the parsimonious formulation (Section~\ref{subsec:parsimonious}) to the critical set $\calCpar$.}
	\label{tab:solver_stats}
	\begin{tabular}{lcccc}
		\toprule
		 & NOM & ROB-S & PAR-S & PAR-SP \\
		\midrule
		\multicolumn{5}{l}{\emph{Warm start}} \\
		decision variables & --- & 19362 & 19362 & 29463 \\
		equality / inequality constr. & --- & 18900 / 700 & 18900 / 700 & 28980 / 700 \\
		IPOPT iterations & --- & 62 & 62 & 81 \\
		wall time (s) & --- & 24.2 & 24.2 & 44.6 \\
		\midrule
		\multicolumn{5}{l}{\emph{Final solve}} \\
		decision variables & 4203 & 34023 & 15528 & 25728 \\
		equality / inequality constr. & 3780 / 700 & 33600 / 840 & 15069 / 840 & 25245 / 840 \\
		IPOPT iterations & 113 & 69 & 66 & 105 \\
		wall time (s) & 28.1 & 227.2 & 85.6 & 590.0 \\
		\midrule
		total wall time (s) & 28.1 & 251.3 & 109.8 & 634.6 \\
		nodes with robust constraints & 0\% & 100\% & 34.3\% & 34.3\% \\
		\bottomrule
	\end{tabular}
\end{table}

Returning to Figure~\ref{fig:adherence_ref}, we can now observe the effect of the robust friction-limit constraints in ROB-S, PAR-S and PAR-SP: when the vehicle negotiates the low-speed corner at $\alpha \in [0.71, 0.73]$, NOM reaches full saturation on the front and rear axles, while the three robust references preserve a margin from the saturation bound. To achieve this, all the robust MLTPs begin braking slightly earlier than NOM ahead of the corner entry, at $\alpha \in [0.70, 0.71]$. The four references coincide for $\alpha \in [0.735, 0.77]$, where the reduced tyre usage leaves the friction-limit constraint inactive.
Interestingly, ROB-S and PAR-S show identical saturation profiles, yet PAR-S requires only $109.8$\,s against the $251.3$\,s of ROB-S (Table~\ref{tab:solver_stats}): the parsimonious formulation preserves solution accuracy at less than half the computational cost.
Finally, PAR-SP introduces a slightly larger margin from full saturation than ROB-S and PAR-S, consistent with the additional parameter uncertainty it accounts for.

\subsection{MPC design} \label{subsec:mpc_design}

Using an MPC as the virtual driver, rather than a human in a driving simulator, makes the closed-loop trials reproducible and allows a statistically meaningful number of runs.
The MPC internal model is the same Single-Track model used in the MLTP (Section~\ref{subsec:vehicle_model}), with state $\bx\in\bbR^{n_x}$ and control $\bu\in\bbR^{n_u}$ of dimensions $n_x=6$ and $n_u=2$; the uncertain vehicle parameters are held at their nominal values (Table~\ref{tab:vehicle_params}).
At each sampling instant~$k$, the MPC predicts the vehicle trajectory and plans an optimal control sequence over a horizon of $N_{\mathrm{pred}}=10$ stages spaced by $h=0.01$~s.

The pairs $(\bx_{k,i},\bu_{k,i})$ of predicted states and planned controls---the first subscript $k$ referring to the current sampling instant, the second $i=0,\ldots,N_{\mathrm{pred}}$ to the prediction stage---constitute the set of decision variables of the MPC, while their target counterparts are the MLTP reference pairs $(\bx_{k,i}^{\mathrm{ref}},\bu_{k,i}^{\mathrm{ref}})$. 
To track the reference, the MPC solves an OCP with the quadratic cost
\begin{equation}
	\sum_{i=0}^{N_{\mathrm{pred}}-1} \Big( \|\bx_{k,i}-\bx_{k,i}^{\mathrm{ref}}\|_Q^2 + \|\bu_{k,i}-\bu_{k,i}^{\mathrm{ref}}\|_R^2 \Big) + \|\bx_{k,N_{\mathrm{pred}}}-\bx_{k,N_{\mathrm{pred}}}^{\mathrm{ref}}\|_{Q_N}^2,
	\label{eq:mpc_cost}
\end{equation}
where $Q$, $R$ and $Q_N$ weight matrices. 
The OCP is subject to the internal-model dynamics and to the nominal friction-limit constraints~\eqref{eq:friction_limit} enforced along the horizon.
Following the receding-horizon principle, only the first optimal control $\bu_{k,0}$ is applied to the simulated FSAE vehicle before the problem is solved again at~$k+1$ with updated state feedback and reference window.
The problem is solved online with the Advanced-Step Real-Time Iteration (AS-RTI) scheme~\cite{Nurkanovic:AdvancedStepReal:2019,Frey:AdvancedStepRealtimeIterations:2024}---an extension of the real-time iteration approach~\cite{Diehl:NominalStabilityRealtime:2005}---in the \texttt{acados} framework~\cite{Verschueren:AcadosModularOpensource:2022}, performing a single warm-started iteration per instant rather than minimising~\eqref{eq:mpc_cost} to convergence. 
This real-time-capable scheme would allow the same controller to be deployed on an actual vehicle in future work.

\begin{figure}
	\centering
	\input{Fig/TikZ_Sources/tikz_mpc_reference_construction.tex}
	\caption{Schematic of MPC reference generation at sampling instant~$k$, drawn with $N_{\mathrm{pred}}=3$ for clarity ($N_{\mathrm{pred}}=10$ in the experiments).
	The curvilinear abscissa~$\al_k$ of the simulated FSAE car (black car) fixes the horizon origin and yields the reference time $\tilde{t}_k$, which locates the ghost car being tracked (red car).
	The targets $\bx^{\mathrm{ref}}_{k,i} = \bx^{\mathrm{ref}}(\tilde{t}_k + i h)$ are sampled forwards along the prerecorded MLTP reference (red markers and dashed red line, times below the track), while the MPC predicts the states $\bx_{k,i}$ over the same horizon (black markers and dashed black line).}
	\label{fig:mpc_ref_sampling}
\end{figure}
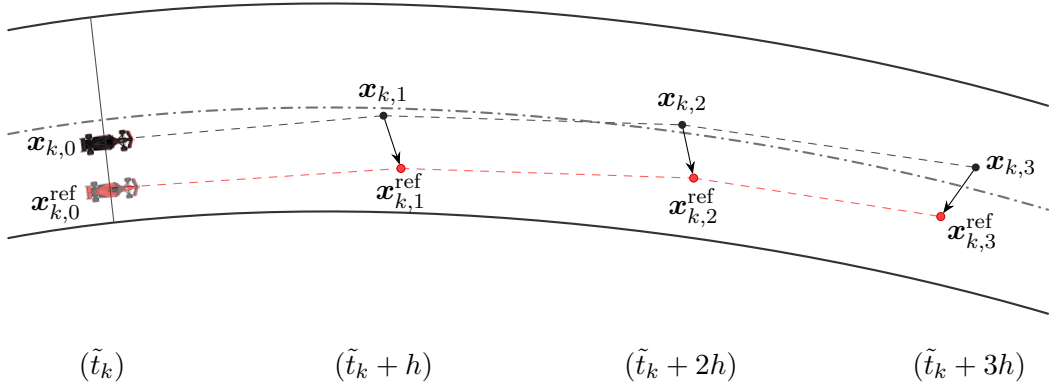

The core of the tracking scheme resides in how we assemble the targets $(\bx_{k,i}^{\mathrm{ref}},\bu_{k,i}^{\mathrm{ref}})$ online from the MLTP reference (Section~\ref{subsec:reference_planning}), as shown in Figure~\ref{fig:mpc_ref_sampling}. 
At sampling instant~$k$, the vehicle's abscissa $\al_k$ along the track synchronises the MPC horizon (dashed black lines) with the reference (dashed red lines): it locates the leading target $(\bx_{k,0}^{\mathrm{ref}},\bu_{k,0}^{\mathrm{ref}})$ and its reference time $\tilde{t}_k$.
From $\tilde{t}_k$ we sample forwards the subsequent targets at the stage temporal spacing~$h$,
\begin{equation}
	\bx^{\mathrm{ref}}_{k,i} = \bx^{\mathrm{ref}}(\tilde{t}_k + i h), \qquad i = 0, \ldots, N_{\mathrm{pred}},
	\label{eq:mpc_ref_sampling}
\end{equation}
and apply the same scheme for the controls.
In summary, in Figure~\ref{fig:mpc_ref_sampling} the black car is the simulated FSAE vehicle, while the red vehicle is the ghost car being followed, corresponding to the MLTP reference. 
In our experiments the MPC does not search for a new optimal trajectory; it faithfully tracks the MLTP reference planned offline, which is the object of validation.

\subsection{Testing strategy} \label{subsec:testing_strategy}

We assess each reference with a Monte Carlo campaign of $1000$ runs, the MPC of Section~\ref{subsec:mpc_design} driving the simulated FSAE vehicle along the planned reference (Section~\ref{subsec:reference_planning}).
The same controller drives all four references, so outcome differences reflect reference driveability alone.
Each run applies three kinds of perturbations mirroring the uncertainty of the robust framework (Section~\ref{sec:robust_mltp}): (i)~a force--moment pulse realising the state uncertainty, (ii)~a random scatter of the vehicle parameters, and (iii)~the additive process noise $\bw$.
The \emph{paired-sample design} with a common seed makes the $i$-th run of every batch face identical disturbances.

The impulse (i) mirrors the model's state uncertainty: we take the current state as the centre of a distribution with covariance $\bP_0$ (the MLTP design values), sample a random state jump, and inject the longitudinal/lateral force and yaw-moment pulses that produce it.
We apply the pulses at $\alpha_P=0.725$, midway through the low-speed corner, where both axles run close to the friction limit in the NOM reference (Figure~\ref{fig:adherence_ref}).
The pulses have duration $T_p=0.1$\,s, are centred at instant $t_p$, and share a raised-cosine profile
\begin{equation}
1+\cos\left[ \frac{2\pi(t-t_p)}{T_p}\right],
\qquad t\in\left[t_p-\frac{T_p}{2},\;t_p+\frac{T_p}{2}\right],
\label{eq:impulse_shape}
\end{equation}
scaled on each of its longitudinal-force, lateral-force and yaw-moment components so that the delivered impulse matches the target state change~\cite{Mastinu:GlobalStabilityRoad:2023}.
The parameter scatter (ii) draws the yaw inertia, centre-of-mass position and aerodynamic drag coefficient (Section~\ref{subsec:vehicle_model}) of the simulated vehicle from a Gaussian about their nominal values, with the standard deviations in the parameter block of $\bP_0$ (Table~\ref{tab:P0}). The sampled parameters are held fixed along the run while the MPC internal model keeps the nominal values; only PAR-SP plans against this uncertainty, yet all four references are tested against it.
The process noise (iii) draws $\bw(t_k)$ at every step from the planning covariance $\bQ$ of the stochastic dynamics~\eqref{eq:dynamics}.

A run is classified as physically failed if any plausibility bound is violated: body sideslip greater than $0.5\,\mathrm{rad}$, yaw rate greater than $3.0\,\mathrm{rad/s}$, or either-axle slip angle greater than $0.3\,\mathrm{rad}$. 
A run that does not trigger this physical-failure criterion reaches the end of the track sector and is therefore classified as completed.
A completed run is further classified as survived if the saturation ratio $S_j$ never exceeds $0.999$ on either axle for a continuous time window longer than the dwell threshold $T_{\mathrm{thr}}$. 
Hence, for each reference, the batch of $1000$ runs gives rise to a completed cohort and a survived subcohort. In Section~\ref{sec:results}, we analyse how $T_{\mathrm{thr}}$ influences the resulting survival counts, i.e., the number of runs in the survived subcohort.

%% file: Fig/TikZ_Sources/fig_ref_adherence.tex
\begin{tikzpicture}
\begin{groupplot}[
  group style={group size=1 by 2, vertical sep=4mm, x descriptions at=edge bottom},
  width=\dispwidth, height=\dispheight,
  xmin=0.70, xmax=0.77,
  ymin=0.0,  ymax=1.1,
  xtick={0.70,0.71,0.72,0.73,0.74,0.75,0.76,0.77},
  ytick={0,0.2,0.4,0.6,0.8,1.0},
  enlarge x limits=false,
  tick align=outside, tick pos=left,
  grid=major, grid style={gray!30, line width=0.2pt},
  every axis plot/.append style={no marks, line width=0.8pt},
]
\nextgroupplot[ylabel={$S_1$},
  legend cell align=left,
  legend style={font=\scriptsize, draw=none, fill=white, fill opacity=0.85,
                text opacity=1, at={(0.97,0.97)}, anchor=north east},
]
  \addplot[nomcolor]
    table[col sep=comma, x expr=\thisrow{s}/4647, y=sat_front]
    {post_proc/ref_solutions_csv/nominal_1_short_adherence_reference.csv};
  \addlegendentry{NOM}
  \addplot[robscolor]
    table[col sep=comma, x expr=\thisrow{s}/4647, y=sat_front]
    {post_proc/ref_solutions_csv/robust_s_4_short_adherence_reference.csv};
  \addlegendentry{ROB-S}
  \addplot[parscolor, dashed]
    table[col sep=comma, x expr=\thisrow{s}/4647, y=sat_front]
    {post_proc/ref_solutions_csv/lazy_s_4_short_adherence_reference.csv};
  \addlegendentry{PAR-S}
  \addplot[parspcolor]
    table[col sep=comma, x expr=\thisrow{s}/4647, y=sat_front]
    {post_proc/ref_solutions_csv/lazy_s_p_4_short_adherence_reference.csv};
  \addlegendentry{PAR-SP}
  \addplot[gray, densely dotted, line width=0.8pt, forget plot,
           domain=0.70:0.77, samples=2] {1};
\nextgroupplot[ylabel={$S_2$}, xlabel={$\alpha$}]
  \addplot[nomcolor]
    table[col sep=comma, x expr=\thisrow{s}/4647, y=sat_rear]
    {post_proc/ref_solutions_csv/nominal_1_short_adherence_reference.csv};
  \addplot[robscolor]
    table[col sep=comma, x expr=\thisrow{s}/4647, y=sat_rear]
    {post_proc/ref_solutions_csv/robust_s_4_short_adherence_reference.csv};
  \addplot[parscolor, dashed]
    table[col sep=comma, x expr=\thisrow{s}/4647, y=sat_rear]
    {post_proc/ref_solutions_csv/lazy_s_4_short_adherence_reference.csv};
  \addplot[parspcolor]
    table[col sep=comma, x expr=\thisrow{s}/4647, y=sat_rear]
    {post_proc/ref_solutions_csv/lazy_s_p_4_short_adherence_reference.csv};
  \addplot[gray, densely dotted, line width=0.8pt, forget plot,
           domain=0.70:0.77, samples=2] {1};
  \node[font=\scriptsize, align=center] at (axis cs:0.725,0.7) {low-speed\\corner};
  \node[font=\scriptsize, align=center] at (axis cs:0.755,0.9) {high-speed\\corner};
\end{groupplot}
\end{tikzpicture}

%% file: Fig/TikZ_Sources/tikz_mpc_reference_construction.tex
\makeatletter
\newcommand{\tikztrackcarimg}{Fig/TikZ_Sources/car_black.png}
\newcommand{\tikztrackcarrefimg}{Fig/TikZ_Sources/car_red.png}
\IfFileExists{car_black.png}{\renewcommand{\tikztrackcarimg}{car_black.png}}{}
\IfFileExists{car_red.png}{\renewcommand{\tikztrackcarrefimg}{car_red.png}}{}
\def\tikztrackcarrefopacity{0.65}
\makeatother
\usetikzlibrary{positioning}
\begin{tikzpicture}[
	x=2.5cm, y=2.5cm,
	trackborder/.style={thick, draw=black!80, line cap=round},
	centerline/.style={
		thick, draw=black!55, line cap=round,
		dash pattern=on 4pt off 2.5pt on 0.4pt off 2.5pt,
	},
	vehpoint/.style={circle, fill=black!85, inner sep=0pt, minimum size=3pt},
	widepoint/.style={circle, fill=red!75, draw=red!85!black, inner sep=0pt, minimum size=3pt},
	projline/.style={thin, draw=black!70, line cap=round},
	projlinewide/.style={thin, draw=red!65, line cap=round},
	harrow/.style={thin, <->, >=Stealth, shorten >=1.5pt},
	bpred/.style={thin, ->, >=Stealth, shorten >=1.5pt},
	directlabel/.style={font=\normalsize},
]

\coordinate (Lstart) at (0, 0.75);
\coordinate (Lend)   at (5.5, 0.35);
\coordinate (Cstart) at (0, 0.20);
\coordinate (Cend)   at (5.5, -0.20);
\coordinate (Rstart) at (0, -0.35);
\coordinate (Rend)   at (5.5, -0.75);

\draw[centerline] (Cstart) .. controls (1.6, 0.50) and (3.8, 0.30) .. (Cend);

\def\latoff{-1.50}
\def\latoffwide{-5.0}
\def\angvehlast{87}
\def\angreflast{93}
\def\tqstart{0.10}
\def\tqspan{0.82}
\def\nsamples{3}
\foreach \i in {0,...,\nsamples} {
	\pgfmathsetmacro{\tq}{\tqstart + \tqspan*\i/\nsamples}
	\pgfmathsetmacro{\tqminus}{max(0.01, \tq - 0.014)}
	\pgfmathsetmacro{\tqplus}{min(0.99, \tq + 0.014)}
	\path
		(Cstart) .. controls (1.6, 0.50) and (3.8, 0.30) .. (Cend)
		coordinate[pos=\tq] (Q\i)
		coordinate[pos=\tqminus] (Qpminus\i)
		coordinate[pos=\tqplus] (Qpplus\i);
	\def\av{90}\def\ar{90}
	\ifnum\i=\nsamples\relax
		\def\av{\angvehlast}\def\ar{\angreflast}
	\fi
	\coordinate (B\i) at ($(Q\i)!{(\latoff+3.0*\i/\nsamples)}!\av:(Qpplus\i)$);
	\coordinate (P\i) at ($(Q\i)!{(\latoffwide+3.0*\i/\nsamples)}!\ar:(Qpplus\i)$);
	\ifnum\i=\nsamples\relax
		\coordinate (Blast) at (B\i);
		\coordinate (Plast) at (P\i);
	\fi
}

\coordinate (P1) at ($(P1) + (0.10, 0)$);
\coordinate (P2) at ($(P2) + (0.10, 0)$);
\coordinate (P3) at ($(P3) - (0.12, 0)$);
\coordinate (Plast) at (P3);

\draw[projline]
	($(B0)!{2.55}!180:(P0)$)
	-- ($(P0)!{-.7}!0:(B0)$);


\foreach \i in {1,...,\nsamples} {
	\pgfmathtruncatemacro{\iprev}{\i-1}
	\draw[projline, dashed] (B\iprev) -- (B\i);
	\draw[projlinewide, dashed] (P\iprev) -- (P\i);
}

\foreach \i in {0,...,\nsamples} {
	\ifnum\i>0\relax
		\draw[bpred] (B\i) -- (P\i);
	\fi
	\ifnum\i>0\relax
		\node[vehpoint] at (B\i) {};
		\node[widepoint] at (P\i) {};
	\fi
}

\path let \p1=(Qpplus0), \p2=(Qpminus0), \n2={atan2(\y1-\y2,\x1-\x2)} in
	node[anchor=center, inner sep=0, rotate=\n2,
		opacity=\tikztrackcarrefopacity] at (P0) {%
		\includegraphics[width=0.75cm, angle=180]{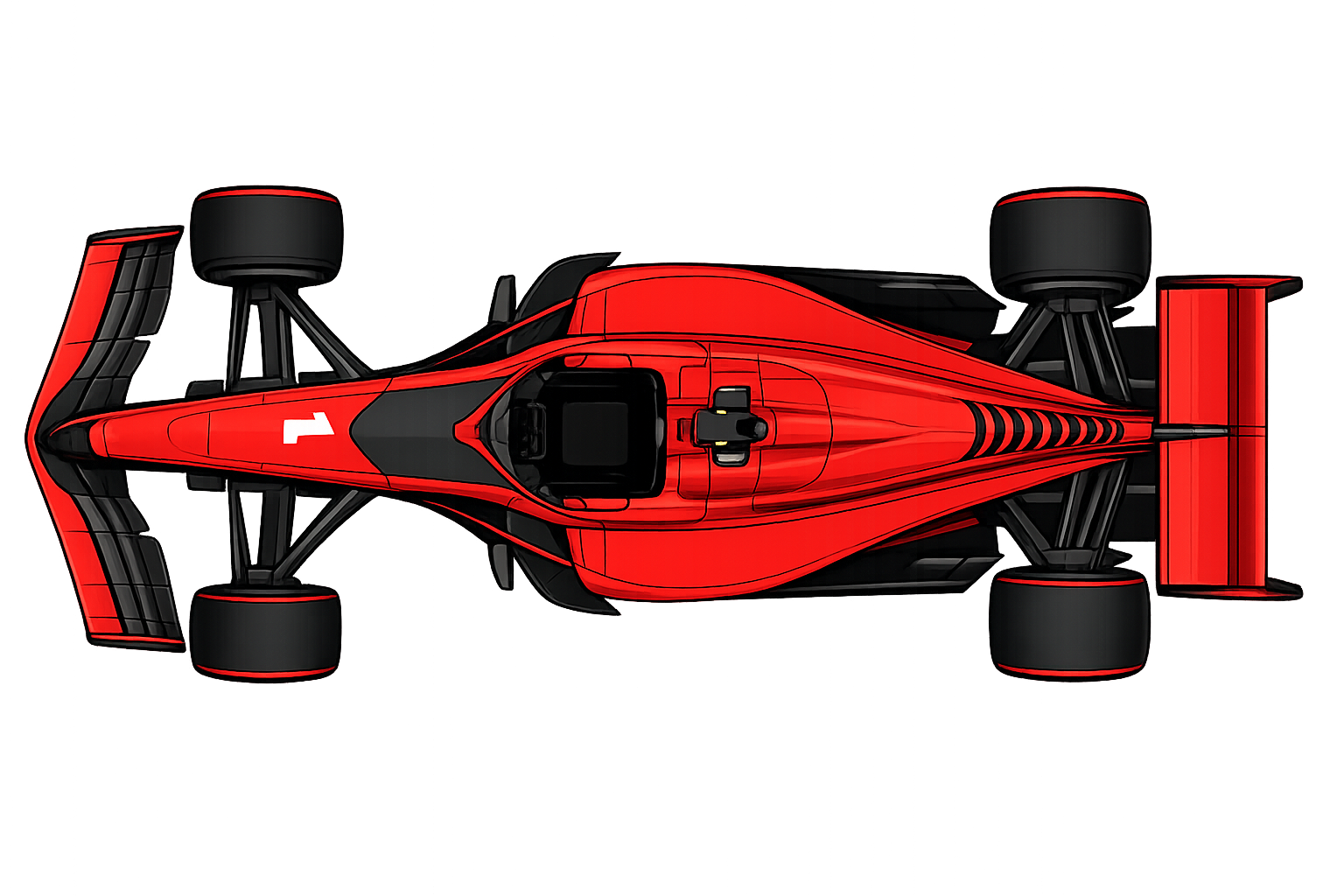}%
	}
	node[anchor=center, inner sep=0, rotate=\n2] at (B0) {%
		\includegraphics[width=0.75cm, angle=180]{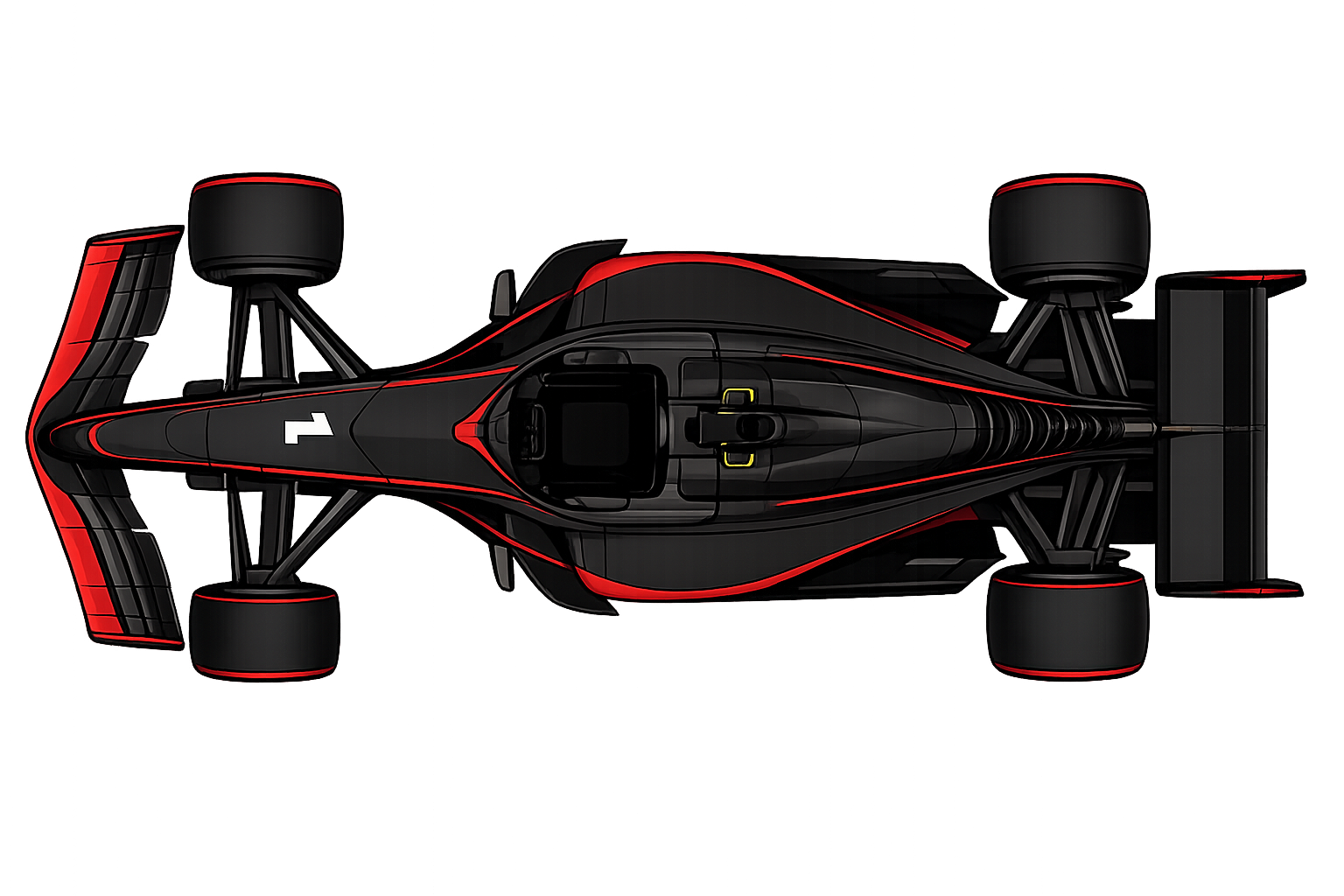}%
	};

\node[directlabel, left = 5pt] at ($(B0)!{0.34}!180:(Qpplus0)$) {$\bx_{k,0}$};
\node[directlabel, below left=-5pt and 7pt of P0] {$\bx^{\mathrm{ref}}_{k,0}$};
\node[directlabel, right=0pt] at (Blast) {$\bx_{k,\nsamples}$};
\node[directlabel, below right=-5pt and -1.5pt of Plast] {$\bx^{\mathrm{ref}}_{k,\nsamples}$};

\node[directlabel, above=0pt] at (B1) {$\bx_{k,1}$};
\node[directlabel, above=0pt of B2] {$\bx_{k,2}$};
\node[directlabel, below=-2pt] at (P1) {$\bx^{\mathrm{ref}}_{k,1}$};
\node[directlabel, below=0pt of P2] {$\bx^{\mathrm{ref}}_{k,2}$};

\draw[trackborder]
	(Lstart) .. controls (1.6, 1.05) and (3.8, 0.85) .. (Lend);

\draw[trackborder]
	(Rstart) .. controls (1.6, -0.05) and (3.8, -0.25) .. (Rend);

\coordinate (timerow) at (0,-1.02);
\foreach \i/\tlab in {%
    0/{\tilde{t}_k},%
    1/{\tilde{t}_k+h},%
    2/{\tilde{t}_k+2h},%
    3/{\tilde{t}_k+3h}%
} {
    \node[directlabel] at (Q\i |- timerow) {$(\tlab)$};
}

\end{tikzpicture}

%% file: results.tex
\section{Results and discussion} \label{sec:results}

\subsection{Driveability analysis} \label{subsec:driveability_analysis}

We assess the driveability of each MLTP reference (Section~\ref{subsec:reference_planning}) by means of the fraction of survived runs (Section~\ref{subsec:testing_strategy}) out of the $1000$ perturbed runs the MPC virtual driver (Section~\ref{subsec:mpc_design}) drives tracking that reference.
The paired campaign applies identical disturbances to all four references, so differences in failure rate are attributable to the references alone.
A run is declared failed based on the criteria introduced in Section~\ref{subsec:testing_strategy}; we record the track position where each failed run first meets one of these criteria.
For completed but failed runs, the failure location corresponds to the end of the $T_{\mathrm{thr}}$ adherence-dwell window.

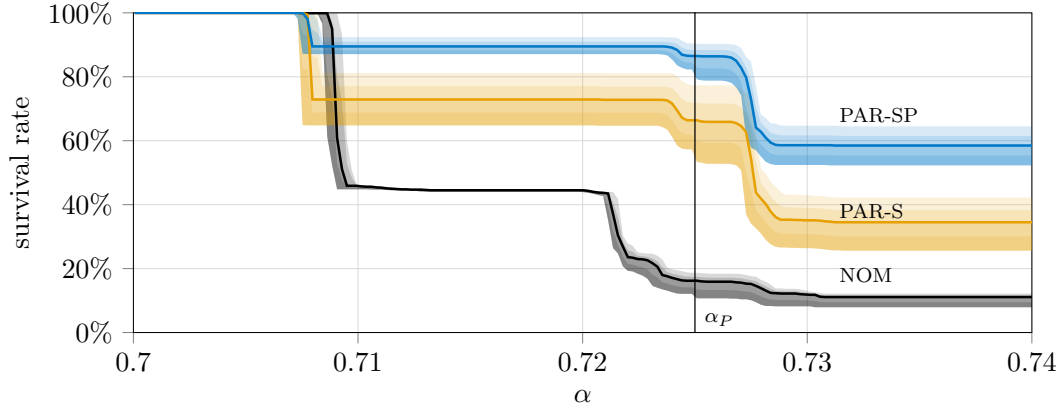
\begin{figure}
	\centering
	\resizebox{\linewidth}{!}{\input{Fig/TikZ_Sources/fig_survival}}
	\caption{Fraction of the survived runs for each 1000 runs batch vs $\alpha$ at a friction-limit dwell threshold $T_{\mathrm{thr}}=0.10$\,s (solid lines), with the shaded band spanning $T_{\mathrm{thr}}\in[0.05,0.15]$\,s. Curves and bands are shown for NOM (black), PAR-S (yellow) and PAR-SP (blue); ROB-S coincides with PAR-S and is omitted. The vertical line marks the impulse abscissa $\alpha_P=0.725$, and each failure is counted at the position where the failure criteria are first met.}
	\label{fig:survival}
\end{figure}

Figure~\ref{fig:survival} reports the fraction of survived runs against the track abscissa $\alpha$, with the vertical line at $\alpha=\alpha_P$ indicating the location of the impulsive disturbance actions. 
For $T_{\mathrm{thr}}=0.10$\,s the sector-end survival counts are NOM $111/1000$ ($11.1\%$, black line), PAR-S $345/1000$ ($34.5\%$, yellow line), and PAR-SP $585/1000$ ($58.5\%$, blue line); the shaded regions show how a different choice of $T_{\mathrm{thr}}$ would affect the survival counts, with the upper bound of each band corresponding to the more permissive $T_{\mathrm{thr}}=0.15$\,s and the lower bound to the stricter threshold $T_{\mathrm{thr}}=0.05$\,s. ROB-S is omitted since its counts coincide with PAR-S.
Completion, by contrast, is nearly universal: ROB-S, PAR-S and PAR-SP complete every one of the $1000$ realisations, while NOM completes $984/1000$, the $16$ shortfall runs leaving their plausibility bound before reaching the sector end. The common completed cohort (Section~\ref{subsec:testing_strategy}) therefore coincides with NOM's own completed set, $984/1000$, and underpins the analyses of Sections~\ref{subsec:driveability_analysis} and~\ref{subsec:time_and_effort}.

\begin{figure}
	\centering
	\subfloat[NOM]{\includegraphics[width=0.48\linewidth]{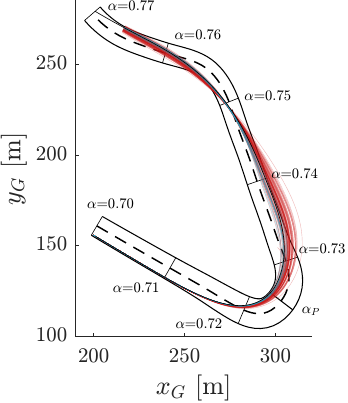}}\hfill
	\subfloat[PAR-SP]{\includegraphics[width=0.48\linewidth]{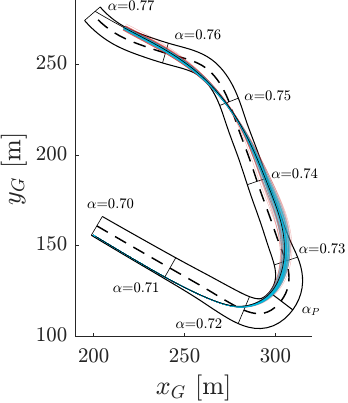}}
	\caption{Representative trajectories for the NOM (a) and PAR-SP (b) references; survived runs in blue, failed runs in red, the MLTP reference in black, and the pulse position marked $\alpha_P$. For each reference, $100$ of its $1000$ runs are drawn in proportion to its survival rate, so the blue-to-red ratio mirrors the counts of Figure~\ref{fig:survival}.}
	\label{fig:traj_samples}
\end{figure}

To visualise the different behaviour of the survived and failed runs, we plot the trajectories of a representative sample of the Monte Carlo campaign for the NOM and PAR-SP (Figure~\ref{fig:traj_samples}(a)--(b)).
Each panel draws $100$ runs independently from that reference's own campaign, in proportion to its end-section survival rate, so the blue (survived) to red (failed) ratio mirrors the counts above.

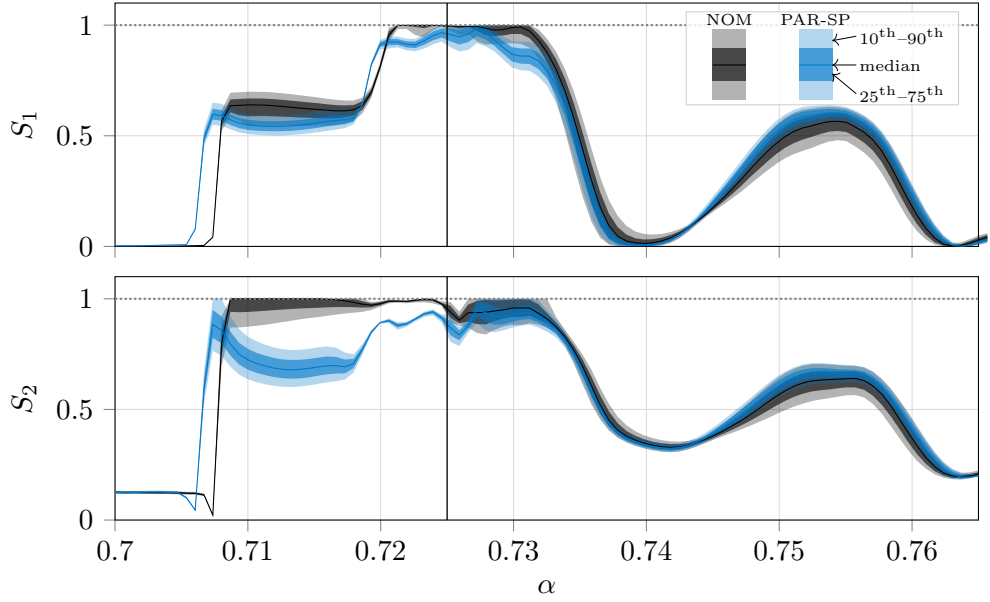
\begin{figure}
	\centering
	\input{Fig/TikZ_Sources/fig_dispersion_nom_parsp}
	\caption{NOM (black) vs PAR-SP (blue): front ($S_1$, top) and rear ($S_2$, bottom) axle saturation ratio vs $\alpha$ on the common completed cohort. Solid line: median; darker band: 25th--75th percentile interval; lighter band: 10th--90th. The dotted line marks the friction limit $S_j=1$; the vertical line marks the impulse abscissa $\alpha_P=0.725$. ROB-S and PAR-S are omitted, their distributions nearly overlapping that of PAR-SP.}
	\label{fig:adherence_dispersion}
\end{figure}

Figure~\ref{fig:adherence_dispersion} illustrates the adherence behaviour of the whole batch relative to each reference.
Here, we analyse the dispersion of the axle saturation indices over the common completed cohort of $984/1000$ runs, so that every band spans the full track sector for the same paired-seed runs. 
At each abscissa $\alpha$ we take the 10th, 25th, 50th, 75th and 90th percentiles of $S_1$ and $S_2$ (\eqref{eq:axle_saturation}) and plot them for NOM (black) and PAR-SP (blue): solid lines are the medians, the darker bands the 25th--75th percentile range, the lighter bands the 10th--90th. For clarity we omit ROB-S and PAR-S, whose distributions nearly overlap that of PAR-SP.

The adherence profiles in Figure~\ref{fig:adherence_dispersion} reflect the MLTP reference curves of Figure~\ref{fig:adherence_ref}, yet the bands reveal how some runs tracking a robust reference may still hit tyre saturation under certain disturbance realisations. 
For instance, although the PAR-SP reference keeps a margin to the friction limit, the 25th--75th percentile band of $S_2$ for PAR-SP reaches saturation over $\alpha\in[0.70,0.71]$ during braking into the slow-speed corner; this explains the survival-rate drop at the same $\alpha$ range in Figure~\ref{fig:survival}. Each remaining survival-rate drop in Figure~\ref{fig:survival} corresponds to one of the percentile bands reaching the friction limit.
More trajectory samples and further telemetry signals are available in the supplementary material~\citep{Gulisano:ParsimoniousDisturbanceAwareData:2026}.

\subsection{Sector-time cost and steering effort} \label{subsec:time_and_effort}

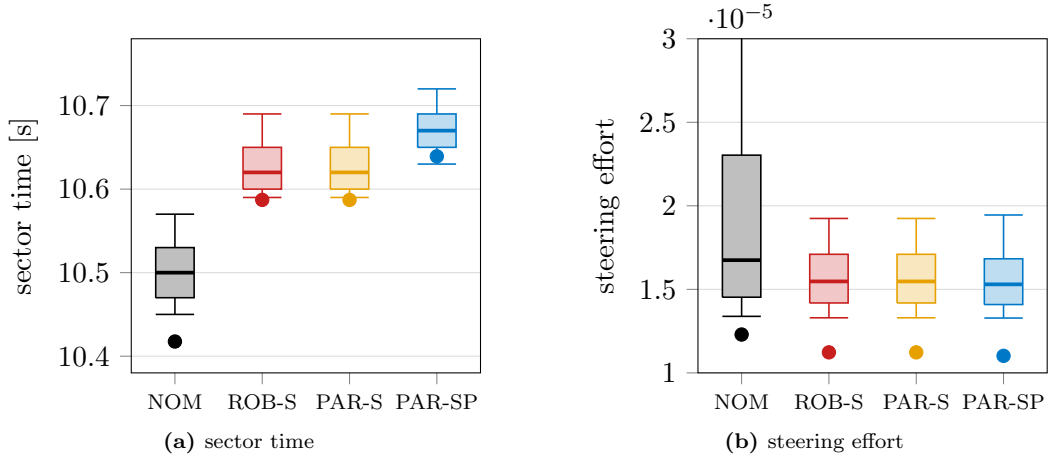
\begin{figure}
	\centering
	\subfloat[sector time]{\input{Fig/TikZ_Sources/fig_sector_time}}\hfill
	\subfloat[steering effort]{\input{Fig/TikZ_Sources/fig_steer_energy}}
	\caption{Distribution of sector time (a) and steering effort (b) for the four references, over all completed runs of each batch (survived and failed alike). Thick line: median; box: 25th--75th percentile interval; whiskers: 10th--90th percentile interval; dot: planned MLTP value. Panel (b) is cropped to $[1,3]\times10^{-5}$; NOM's upper whisker (90th percentile) reaches $6.6\times10^{-5}$, off the chart.}
	\label{fig:time_and_effort}
\end{figure}

Robustness carries a cost: the back-off terms that hold the planned trajectory off the friction boundary entail a higher sector time. 
Figure~\ref{fig:time_and_effort}(a) shows the sector-time distribution over the common completed cohort (Section~\ref{subsec:driveability_analysis}). For each reference, the boxplot includes the median (thick line), the 25th--75th percentile box, the 10th--90th whiskers, while the dot indicates the planned MLTP sector time.
As expected, median sector time rises monotonically with robustness: NOM $10.50$\,s, ROB-S and PAR-S $10.62$\,s ($+120$\,ms over NOM), and PAR-SP $10.67$\,s ($+170$\,ms over NOM, $+50$\,ms over ROB-S/PAR-S).
ROB-S and PAR-S share identical distributions because they plan the same reference (Figure~\ref{fig:adherence_ref}).

On the other hand, the steering effort $\int \dot{\de}^2\,\mathrm{d}t$---a measure of the overall steering activity required to follow the reference---decreases with robustness, as shown in Figure~\ref{fig:time_and_effort}(b). The boxplots describe the steering effort distribution for each reference, while the dot indicates the planned MLTP effort. The plot is cropped to $[1,3]\times10^{-5}$ for readability; NOM's 90th-percentile whisker extends to $6.6\times10^{-5}$, beyond the plotted range.

\subsection{Discussion} \label{subsec:discussion}

The campaign confirms that the robust MLTP framework (Section~\ref{sec:robust_mltp}) produces trajectory and control references that keep the vehicle away from tyre saturation, as Figure~\ref{fig:survival} shows. Although the exact number of survivors depends on the dwell window $T_{\mathrm{thr}}$ over which saturation is tolerated, the shaded band shows a consistent trend: accounting for state uncertainty (ROB-S/PAR-S) raises the survived fraction over NOM, and adding parameter uncertainty (PAR-SP) raises it further still.

The representative samples of Figure~\ref{fig:traj_samples} illustrate this difference. The NOM runs deviate markedly off-line over $\alpha\in[0.73,0.75]$ and spread over a wider corridor than the robust PAR-SP runs, signalling a partial loss of vehicle control and a higher sensitivity to the applied disturbances.

Since the underlying problem is one of minimum-time planning, this robustness has a natural counterpart in the sector time (Figure~\ref{fig:time_and_effort}(a)): moving from NOM to ROB-S/PAR-S costs a median $120$\,ms, while accounting also for parameter uncertainty (PAR-SP) adds a further $50$\,ms. In all four cases the MPC driver does not, on average, reproduce the planned MLTP sector time (dots), settling at higher median values (thick lines); this gap, however, narrows as robustness increases---for PAR-SP, more than $10\%$ of runs beat the planned time, with the MLTP dot lying above the 10th-percentile whisker.

The steering-effort panel (Figure~\ref{fig:time_and_effort}(b)) further measures driveability---understood here as the ease with which a driver reproduces the MLTP reference. The NOM runs require repeated steering corrections that skew their distribution asymmetrically towards high values, with a 90th percentile markedly larger than those of the three robust references, which remain close to one another at a far lower effort.

Although the campaign relies on an MPC virtual driver, the higher driveability of the robust references---greater survival and lower steering effort---plausibly carries over to a human driver, a hypothesis that further testing on a driving simulator or on track could confirm.

%% file: Fig/TikZ_Sources/fig_survival.tex

\newcommand{\gradientband}[2]{%
  \addplot[draw=none, forget plot, name path=gbpath@lo]
    table[col sep=comma, x=\xcol, y expr=\thisrow{#1}/\survN]
    {post_proc/survival_count_vs_s_window_05cent.csv};
  \addplot[draw=none, forget plot, name path=gbpath@q1]
    table[col sep=comma, x=\xcol, y expr=\thisrow{#1}/\survN]
    {post_proc/survival_count_vs_s_window_075cent.csv};
  \addplot[draw=none, forget plot, name path=gbpath@mid]
    table[col sep=comma, x=\xcol, y expr=\thisrow{#1}/\survN]
    {post_proc/survival_count_vs_s_window_10cent.csv};
  \addplot[draw=none, forget plot, name path=gbpath@q3]
    table[col sep=comma, x=\xcol, y expr=\thisrow{#1}/\survN]
    {post_proc/survival_count_vs_s_window_125cent.csv};
  \addplot[draw=none, forget plot, name path=gbpath@hi]
    table[col sep=comma, x=\xcol, y expr=\thisrow{#1}/\survN]
    {post_proc/survival_count_vs_s_window_15cent.csv};
  \addplot[#2, fill opacity=0.15, draw=none, forget plot]
    fill between[of=gbpath@lo and gbpath@hi];
  \addplot[#2, fill opacity=0.15, draw=none, forget plot]
    fill between[of=gbpath@lo and gbpath@q3];
  \addplot[#2, fill opacity=0.15, draw=none, forget plot]
    fill between[of=gbpath@lo and gbpath@mid];
  \addplot[#2, fill opacity=0.15, draw=none, forget plot]
    fill between[of=gbpath@lo and gbpath@q1];
}

\begin{tikzpicture}
\begin{axis}[
  width=14cm, height=6cm,
  xmin=\axxmin, xmax=0.74, ymin=0, ymax=1.0,
  xtick={0.70,0.71,0.72,0.73,0.74},
  ytick={0,0.2,0.4,0.6,0.8,1.0},
  yticklabels={0\%,20\%,40\%,60\%,80\%,100\%},
  enlarge x limits=false,
  tick align=outside, tick pos=left,
  grid=major, grid style={gray!30, line width=0.2pt},
  xlabel={$\alpha$}, ylabel={survival rate},
  every axis plot/.append style={no marks, line width=1pt},
  legend style={draw=none, fill=none},
]
  \gradientband{nominal_1_short}{nomcolor}
  \addplot[nomcolor, forget plot]
    table[col sep=comma, x=\xcol, y expr=\thisrow{nominal_1_short}/\survN]
    {post_proc/survival_count_vs_s_window_10cent.csv};


  \gradientband{lazy_s_4_short}{parscolor}
  \addplot[parscolor, forget plot]
    table[col sep=comma, x=\xcol, y expr=\thisrow{lazy_s_4_short}/\survN]
    {post_proc/survival_count_vs_s_window_10cent.csv};

  \gradientband{lazy_s_p_4_short}{parspcolor}
  \addplot[parspcolor, forget plot]
    table[col sep=comma, x=\xcol, y expr=\thisrow{lazy_s_p_4_short}/\survN]
    {post_proc/survival_count_vs_s_window_10cent.csv};

  \impulseline
  \node[anchor=west, font=\scriptsize, black]
    at (axis cs:\impulsealpha, 0.04) {$\alpha_P$};

  \node[anchor=west, font=\scriptsize, black]
    at (axis cs:0.731, 0.18) {NOM};
  \node[anchor=west, font=\scriptsize, black]
    at (axis cs:0.731, 0.38) {PAR-S};
  \node[anchor=west, font=\scriptsize, black]
    at (axis cs:0.731, 0.68) {PAR-SP};
\end{axis}
\end{tikzpicture}

%% file: Fig/TikZ_Sources/fig_dispersion_nom_parsp.tex
\begin{tikzpicture}
\begin{groupplot}[
  group style={group size=1 by 2, vertical sep=4mm, x descriptions at=edge bottom},
  width=\dispwidth, height=\dispheight,
  xmin=\axxmin, xmax=\axxmax, ymin=\aymin, ymax=\aymax,
  xtick={0.70,0.71,0.72,0.73,0.74,0.75,0.76},
  enlarge x limits=false,
  tick align=outside, tick pos=left,
  every axis plot/.append style={no marks},
  grid=major, grid style={gray!30, line width=0.2pt},
  legend cell align=left, legend columns=1,
  legend style={font=\scriptsize, draw=none, fill=none,
                at={(0.02,0.98)}, anchor=north west},
]
\nextgroupplot[ylabel={$S_1$}, clip=false]
  \plotref{nominal_1_short}{adherence_front}{nomcolor}
  \plotref{lazy_s_p_4_short}{adherence_front}{parspcolor}
  \limitline
  \impulseline
  \bandlegend{nomcolor}{NOM}{parspcolor}{PAR-SP}
\nextgroupplot[ylabel={$S_2$}, xlabel={$\alpha$}]
  \plotref{nominal_1_short}{adherence_rear}{nomcolor}
  \plotref{lazy_s_p_4_short}{adherence_rear}{parspcolor}
  \limitline
  \impulseline
\end{groupplot}
\end{tikzpicture}

%% file: Fig/TikZ_Sources/fig_sector_time.tex
\pgfplotstableread[col sep=comma]{post_proc/sector_time_box_stats.csv}\boxST
\def\hw{0.22}  

\newcommand{\drawbox}[3]{%
  \pgfplotstablegetelem{#3}{p10}\of\boxST    \pgfmathsetmacro\qa{\pgfplotsretval/1000}%
  \pgfplotstablegetelem{#3}{p25}\of\boxST    \pgfmathsetmacro\qb{\pgfplotsretval/1000}%
  \pgfplotstablegetelem{#3}{p50}\of\boxST    \pgfmathsetmacro\qc{\pgfplotsretval/1000}%
  \pgfplotstablegetelem{#3}{p75}\of\boxST    \pgfmathsetmacro\qd{\pgfplotsretval/1000}%
  \pgfplotstablegetelem{#3}{p90}\of\boxST    \pgfmathsetmacro\qe{\pgfplotsretval/1000}%
  \pgfmathsetmacro\xl{#1-\hw}\pgfmathsetmacro\xr{#1+\hw}%
  \addplot[fill=#2!25, draw=#2, line width=0.6pt, forget plot]
    coordinates {(\xl,\qb)(\xr,\qb)(\xr,\qd)(\xl,\qd)(\xl,\qb)};%
  \addplot[#2, line width=1.4pt, forget plot]
    coordinates {(\xl,\qc)(\xr,\qc)};%
  \addplot[#2, line width=0.6pt, forget plot]
    coordinates {(#1,\qa)(#1,\qb)};%
  \addplot[#2, line width=0.6pt, forget plot]
    coordinates {(\xl,\qa)(\xr,\qa)};%
  \addplot[#2, line width=0.6pt, forget plot]
    coordinates {(#1,\qd)(#1,\qe)};%
  \addplot[#2, line width=0.6pt, forget plot]
    coordinates {(\xl,\qe)(\xr,\qe)};%
}

\begin{tikzpicture}
\begin{axis}[
  width=6.2cm, height=6cm,
  xmin=0.5, xmax=4.5,
  ymin=10.38, ymax=10.78,
  ytick={10.4,10.5,10.6,10.7},
  xtick={1,2,3,4}, xticklabels={NOM,ROB-S,PAR-S,PAR-SP},
  ylabel={sector time [s]},
  tick align=outside, tick pos=left,
  ymajorgrids, grid style={gray!30, line width=0.2pt},
  xticklabel style={font=\scriptsize},
  clip=false,
]
  \drawbox{1}{nomcolor}{0}
  \drawbox{2}{robscolor}{1}
  \drawbox{3}{parscolor}{2}
  \drawbox{4}{parspcolor}{3}

  \pgfplotstablegetelem{0}{MLTP_ms}\of\boxST \pgfmathsetmacro\qm{\pgfplotsretval/1000}
  \addplot[only marks,mark=*,mark size=2.5pt,nomcolor,   forget plot] coordinates {(1,\qm)};
  \pgfplotstablegetelem{1}{MLTP_ms}\of\boxST \pgfmathsetmacro\qm{\pgfplotsretval/1000}
  \addplot[only marks,mark=*,mark size=2.5pt,robscolor,  forget plot] coordinates {(2,\qm)};
  \pgfplotstablegetelem{2}{MLTP_ms}\of\boxST \pgfmathsetmacro\qm{\pgfplotsretval/1000}
  \addplot[only marks,mark=*,mark size=2.5pt,parscolor,  forget plot] coordinates {(3,\qm)};
  \pgfplotstablegetelem{3}{MLTP_ms}\of\boxST \pgfmathsetmacro\qm{\pgfplotsretval/1000}
  \addplot[only marks,mark=*,mark size=2.5pt,parspcolor, forget plot] coordinates {(4,\qm)};

\end{axis}
\end{tikzpicture}

%% file: Fig/TikZ_Sources/fig_steer_energy.tex
\pgfplotstableread[col sep=comma]{post_proc/steer_energy_box_stats.csv}\boxSE
\def\hwe{0.22}  

\newcommand{\drawboxE}[3]{%
  \pgfplotstablegetelem{#3}{p10}\of\boxSE  \edef\qa{\pgfplotsretval}%
  \pgfplotstablegetelem{#3}{p25}\of\boxSE  \edef\qb{\pgfplotsretval}%
  \pgfplotstablegetelem{#3}{p50}\of\boxSE  \edef\qc{\pgfplotsretval}%
  \pgfplotstablegetelem{#3}{p75}\of\boxSE  \edef\qd{\pgfplotsretval}%
  \pgfplotstablegetelem{#3}{p90}\of\boxSE  \edef\qe{\pgfplotsretval}%
  \pgfmathsetmacro\xl{#1-\hwe}\pgfmathsetmacro\xr{#1+\hwe}%
  \addplot[fill=#2!25, draw=#2, line width=0.6pt, forget plot]
    coordinates {(\xl,\qb)(\xr,\qb)(\xr,\qd)(\xl,\qd)(\xl,\qb)};%
  \addplot[#2, line width=1.4pt, forget plot]
    coordinates {(\xl,\qc)(\xr,\qc)};%
  \addplot[#2, line width=0.6pt, forget plot]
    coordinates {(#1,\qa)(#1,\qb)};%
  \addplot[#2, line width=0.6pt, forget plot]
    coordinates {(\xl,\qa)(\xr,\qa)};%
  \addplot[#2, line width=0.6pt, forget plot]
    coordinates {(#1,\qd)(#1,\qe)};%
  \addplot[#2, line width=0.6pt, forget plot]
    coordinates {(\xl,\qe)(\xr,\qe)};%
}

\begin{tikzpicture}
\begin{axis}[
  width=6.2cm, height=6cm,
  xmin=0.5, xmax=4.5,
  xtick={1,2,3,4}, xticklabels={NOM,ROB-S,PAR-S,PAR-SP},
  ymin=1e-5, ymax=3e-5,
  ylabel={steering effort},
  tick align=outside, tick pos=left,
  ymajorgrids, grid style={gray!30, line width=0.2pt},
  xticklabel style={font=\scriptsize},
  clip=true,
]
  \drawboxE{1}{nomcolor}{0}
  \drawboxE{2}{robscolor}{1}
  \drawboxE{3}{parscolor}{2}
  \drawboxE{4}{parspcolor}{3}

  \pgfplotstablegetelem{0}{MLTP}\of\boxSE \edef\qm{\pgfplotsretval}
  \addplot[only marks,mark=*,mark size=2.5pt,nomcolor,   forget plot] coordinates {(1,\qm)};
  \pgfplotstablegetelem{1}{MLTP}\of\boxSE \edef\qm{\pgfplotsretval}
  \addplot[only marks,mark=*,mark size=2.5pt,robscolor,  forget plot] coordinates {(2,\qm)};
  \pgfplotstablegetelem{2}{MLTP}\of\boxSE \edef\qm{\pgfplotsretval}
  \addplot[only marks,mark=*,mark size=2.5pt,parscolor,  forget plot] coordinates {(3,\qm)};
  \pgfplotstablegetelem{3}{MLTP}\of\boxSE \edef\qm{\pgfplotsretval}
  \addplot[only marks,mark=*,mark size=2.5pt,parspcolor, forget plot] coordinates {(4,\qm)};

\end{axis}
\end{tikzpicture}

%% file: conclusions.tex
\section{Conclusion} \label{sec:conclusions}

The minimum-time racing line is, by construction, fragile: it runs along the friction-limit boundary with no margin to spare, so a small disturbance, be it a gust, a kerb strike, a momentary loss of grip or a drift in the vehicle parameters, is enough to make the nominal MLTP solution hard or unsafe to follow.
To plan a reference that keeps a conservative margin exactly where it is needed, we presented and validated a parsimonious disturbance-aware framework for minimum-lap-time planning under parametric uncertainty. It extends a prior disturbance-aware MLTP framework along three lines: parameter uncertainty in the covariance propagation, a parsimonious activation strategy for the robust constraints, and validation of the planned references by an MPC virtual driver in a Monte Carlo campaign.

Targeting tyre saturation as the critical condition, we enforced robust friction-limit constraints in the OCP and confined them, through a spatially selective activation strategy, to the circuit segments where they matter most (only $34.3\%$ of the track grid nodes in the tested case). This parsimony costs no accuracy: under state uncertainty alone, the parsimonious reference (PAR-S) reproduces the same trajectory and controls as its full counterpart (ROB-S), in which the robust constraints hold at every node, yet PAR-S solves in $109.8$\,s against $251.3$\,s. This efficiency keeps the formulation tractable once parameter uncertainty enters (PAR-SP).

Validation employed MPC as a virtual test driver: for each of the four planned references (NOM, ROB-S, PAR-S, PAR-SP), the same controller drove a simulated FSAE vehicle over 1000 runs on the Catalunya sector, under randomly drawn impulsive disturbances, scattered vehicle parameters and process noise. With a run counted as failed once it remains at the friction limit longer than the dwell threshold $T_{\mathrm{thr}}=0.10$\,s, the survived fraction rose from $11.1\%$ (NOM) to $34.5\%$ (ROB-S/PAR-S) and $58.5\%$ (PAR-SP), a roughly fivefold safety gain for the full state-and-parameter formulation. The cost of this safety was small: a median of $120$\,ms (ROB-S/PAR-S) and $170$\,ms (PAR-SP) over a $10.50$\,s nominal sector, alongside tighter trajectory dispersion and lower steering effort.

These results rest on a single-track vehicle model, one track sector and a single disturbance location, with the MPC standing in for the driver; within these bounds, the higher driveability of the robust references (greater survival, tighter dispersion, lower steering effort) plausibly carries over to a human driver. The natural next steps are to confirm this on a driving simulator or on track, and to extend the framework to more detailed vehicle models while containing its computational cost.